\documentclass[aapm,mph,notitlepage,graphicx,longbibliography]{revtex4-1}
\usepackage{amsmath} 
\usepackage{graphicx}
\usepackage[table,xcdraw]{xcolor}
\usepackage{tabularx}
\usepackage{subfig}
\usepackage{multirow}
\usepackage{booktabs}
\usepackage{color}
\usepackage{caption}
\usepackage{hhline}
\usepackage{tabularx}
\newcolumntype{L}[1]{>{\raggedright\let\newline\\\arraybackslash\hspace{0pt}}m{#1}}
\newcolumntype{C}[1]{>{\centering\let\newline\\\arraybackslash\hspace{0pt}}m{#1}}
\newcolumntype{R}[1]{>{\raggedleft\let\newline\\\arraybackslash\hspace{0pt}}m{#1}}

\newcommand{\vold}[1]{$#1\!\times\!#1\!\times\!#1$}

\graphicspath{ {Images/}}

\begin{document}


\title{A 3D fully convolutional neural network and a random walker to segment the esophagus in CT} 



\author{Tobias Fechter}
\email[Corresponding Author: ]{tobias.fechter@uniklinik-freiburg.de}
\affiliation{Division of Medical Physics, Department of Radiation Oncology, Medical Center - University of Freiburg, Freiburg, Germany}
\affiliation{German Cancer Consortium (DKTK), Partner Site Freiburg, Germany}
\affiliation{German cancer Research Center (DKFZ), Heidelberg, Germany)}

\author{Sonja Adebahr}
\affiliation{German Cancer Consortium (DKTK), Partner Site Freiburg, Germany}
\affiliation{German cancer Research Center (DKFZ), Heidelberg, Germany)}
\affiliation{Department of Radiation Oncology, Medical Center - University of Freiburg, Faculty of Medicine, University of Freiburg, Germany.}

\author{Dimos Baltas}
\affiliation{Division of Medical Physics, Department of Radiation Oncology, Medical Center - University of Freiburg, Freiburg, Germany}
\affiliation{German Cancer Consortium (DKTK), Partner Site Freiburg, Germany}
\affiliation{German cancer Research Center (DKFZ), Heidelberg, Germany)}

\author{Ismail Ben Ayed}
\affiliation{Laboratory for Imagery, Vision and Artificial Intelligence (LIVIA) \'{E}cole de technologie sup\'{e}rieure, Montr\'{e}al, Canada.}

\author{Christian Desrosiers}
\affiliation{Laboratory for Imagery, Vision and Artificial Intelligence (LIVIA) \'{E}cole de technologie sup\'{e}rieure, Montr\'{e}al, Canada.}

\author{Jose Dolz}
\affiliation{Laboratory for Imagery, Vision and Artificial Intelligence (LIVIA) \'{E}cole de technologie sup\'{e}rieure, Montr\'{e}al, Canada.}

\date{\today}

\begin{abstract}

\textbf{Purpose:} Precise delineation of organs at risk is a crucial task in radiotherapy treatment planning when aiming at delivering high dose to the tumour while sparing healthy tissues. In recent years algorithms showed high performance and the possibility to automate this task for many organs. However, for some organs precise delineation remains challenging, even for human experts. One of them is the esophagus with a versatile shape and poor contrast to neighboring tissue. To tackle these issues we propose a 3D fully CNN (convolutional neural network) driven random walk approach to automatically segment the esophagus on CT images.
\\
\textbf{Methods:} First, a soft probability map is generated by the CNN. Then an active contour model (ACM) is fitted on the CNN soft probability map to get a first estimation of the esophagus location. The outputs of the CNN and ACM are then used in addition to a probability model based on CT Hounsfield (HU) values to drive the random walker. Evaluation and training was done on two different datasets, with a total of 50 CTs with clinically used peer reviewed esophagus contours. Results were assessed regarding spatial overlap and shape similarities.
\\
\textbf{Results:} The esophagus contours, generated by the proposed algorithm showed a mean Dice coefficient of 0.76 $\pm$ 0.11, an average symmetric square distance of 1.36 $\pm$ 0.90 mm and an average Hausdorff distance of 11.68 $\pm$ 6.80 compared to the reference contours. These figures translate into a very good agreement with the reference contours and an increase in accuracy compared to other methods.


\textbf{Conclusion:} We show that by employing a CNN accurate estimations of esophagus location can be obtained and refined by a post processing random walk step taking pixel intensities and neighborhood relationships into account. One of the main advantages compared to previous methods is that our network performs convolutions in a 3D manner, fully exploiting the 3D spatial context and performing an efficient and precise volume-wise prediction. The whole segmentation process is fully automatic and yields esophagus delineations in very good agreement with the used gold standard, showing that it can compete with previously published methods. The results demonstrate the feasibility of our approach employing a CNN to drive a random walker for esophagus segmentation.
\end{abstract}



\maketitle 

\section{Introduction}
%
Precise delineation of organs at risk (OAR) is a crucial task in radiotherapy treatment planning (RTP), which aims at delivering high dose to the tumour while sparing healthy tissues. While some OAR presenting high contrast to neighbouring structures are easy to depict in CT scans, the esophagus is sometimes very difficult to demarcate from the mediastinal structures. On some CT slices even experts have difficulties to define reliably its boundaries and it can easily be confused with other structures. This leads to a tedious interpretation of the CT images, which makes the process not only time-consuming, but also highly prone to inter-observer variability \cite{collier2003}. However, with its radio-sensitive mucosa the esophagus is one of the most critical OAR in RTP, and a reliable contouring is indispensable. Thus, automatic, reproducible and reliable segmentation approaches of the esophagus are greatly desired to be adopted in the RTP.

Although considerable attention has been devoted to automating the process in some other structures in CT \cite{heimann2009statistical}, literature on esophagus segmentation remains very rare with results far from being fully satisfactory. Indeed, presented approaches which successfully segmented a variety of structures on CT often failed to segment the esophagus \cite{ragan2005semiautomated,dolz2016interactive}. One of the main reasons is the absence of consistent intensity contrast between the esophagus and surrounding tissue in thoracic CT scans. Additionally, its appearance varies depending on whether it is filled with air, remains of orally given contrast agent, or both. Furthermore, the esophagus shows a certain mobility. Thus, the esophagus can be considered as a structure with very inhomogeneus appearance and complex shape. Even though some attempts to segment the esophagus have been proposed, most present three main drawbacks: first, they require several levels of user interactivity (placing manual points \cite{rousson2006probabilistic,grosgeorge2013esophagus}, manual region selection \cite{feulner2009fast} or drawing contours at some slices \cite{fieselmann2008esophagus}, for example); second, some of them assume that contours of surrounding organs are available \cite{rousson2006probabilistic,kurugol2011centerline}; and third, registration steps are required in most cases, which may introduce an additional source of error. Despite all these efforts, the automation of this task still remains very challenging, achieving low performance in most cases.  

On the other hand, deep learning has recently emerged as a powerful classification tool, achieving state-of-the art performance in numerous applications of pattern or speech recognition. Among the different types of deep learning approaches, convolutional neural networks (CNNs) \cite{lecun1998gradient} have shown the greatest potential for computer vision and image analysis problems. Particularly in medical image segmentation, CNNs are becoming very popular, achieving outstanding performance in various applications \cite{litjens2017survey}. These architectures are supervised models that are trained end-to-end to learn a hierarchy of features representing different levels of abstraction. In contrast to separate hand-crafted features and classifier models, CNNs can learn simultaneously image features and the classifier. This type of architectures are typically made up of multiple convolution, pooling and fully-connected layers, which parameters are learned using back-propagation by minimizing a given cost function.

The pixel-wise segmentation classes in standard CNN approaches are predicted independently, based on its enclosing region, which generates non-spatial outputs. By reinterpreting classification layers as fully convolutional layers, a CNN network is able to generate spatial output maps for all the classes, becoming a natural choice for dense problems, such as semantic segmentation. These architectures, which are known as fully convolutional neural networks (FCNNs) \cite{long2015fully}, can be seen as a large non-linear filter whose output yields class probabilities. Therefore these networks can accommodate images of arbitrary size and provide much greater efficiency by avoiding redundant convolutions/pooling operations. However, although medical images are commonly presented as 3D images, most of the existing CNN approaches perform the segmentation in a slice-by-slice fashion, not fully exploiting 3D context. Even though the extension of 2D convolutions to 3D seems natural, the additional computational complexity introduces significant challenges, making that 3D CNNs have been mostly avoided up to recent times. Nevertheless, an important benefit of 3D convolutions over 2D representations is their ability to fully exploit dense inference \cite{szegedy2015going}, which allows to decreasing inference times. Recently, 3D FCNNs yielded outstanding segmentation performances in the context of brain lesions \cite{kamnitsas2016efficient} and subcortical brain structures \cite{dolz20163d}. 

Nevertheless, if there is a lack of annotated massive data and/or a high complexity of the problem at hand, CNN architectures tend to produce imprecise segmentations. To further improve segmentation results, some authors have interpreted the CNN output as potentials of a Markov Random Field \cite{shakeri2016sub} or as initialization to drive a level set segmentation \cite{cha2016urinary}, for example. In this way, coarse pixel (or voxel) level label predictions from the CNN are refined to produce sharp boundaries and fine-grained segmentations.

\vspace{5mm}
\paragraph*{\textbf{Contributions:}}

In this study, we explored the feasibility of guiding a random walk (RW) algorithm \cite{gradyML} with a CNN and its application to CT esophagus segmentation. Particularly, we trained a CNN to generate an esophagus probability map, which was used to drive a RW segmentation. We believe this is the first work to address esophagus segmentation by employing deep learning. Additionally, and to the best of our knowledge, this work also represents the first attempt to drive a RW by employing CNN-based probability maps. To generate these probability maps, we employed a 3D fully CNN, which use in the medical field has been very limited up to date. With the use of 3D context, our method can handle volumetric homogeneity better than 2D-based convolutions, which is particularly important in thin tubular structures. Furthermore, most of the existing works address the problem by requiring some sort of user interaction. Contrariwise, our proposed method performs the task in a fully automated manner. And last, we report state-of-the-art performance in one public and in one clinical dataset, showing a good agreement with clinically used peer reviewed esophagus annotations.

%

\vspace{-3mm}
\section{Materials and Methods}
\vspace{-2mm}

\begin{figure}[t!]
     \begin{center}
     \mbox{
      \shortstack{
        \includegraphics[width=0.45\linewidth]{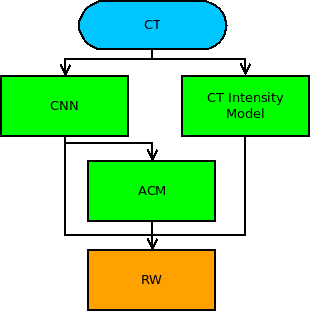} \\
        a) Algorithm structure
        }
        \hspace{2 mm}
         
      \shortstack{     
        \includegraphics[width=0.49\linewidth]{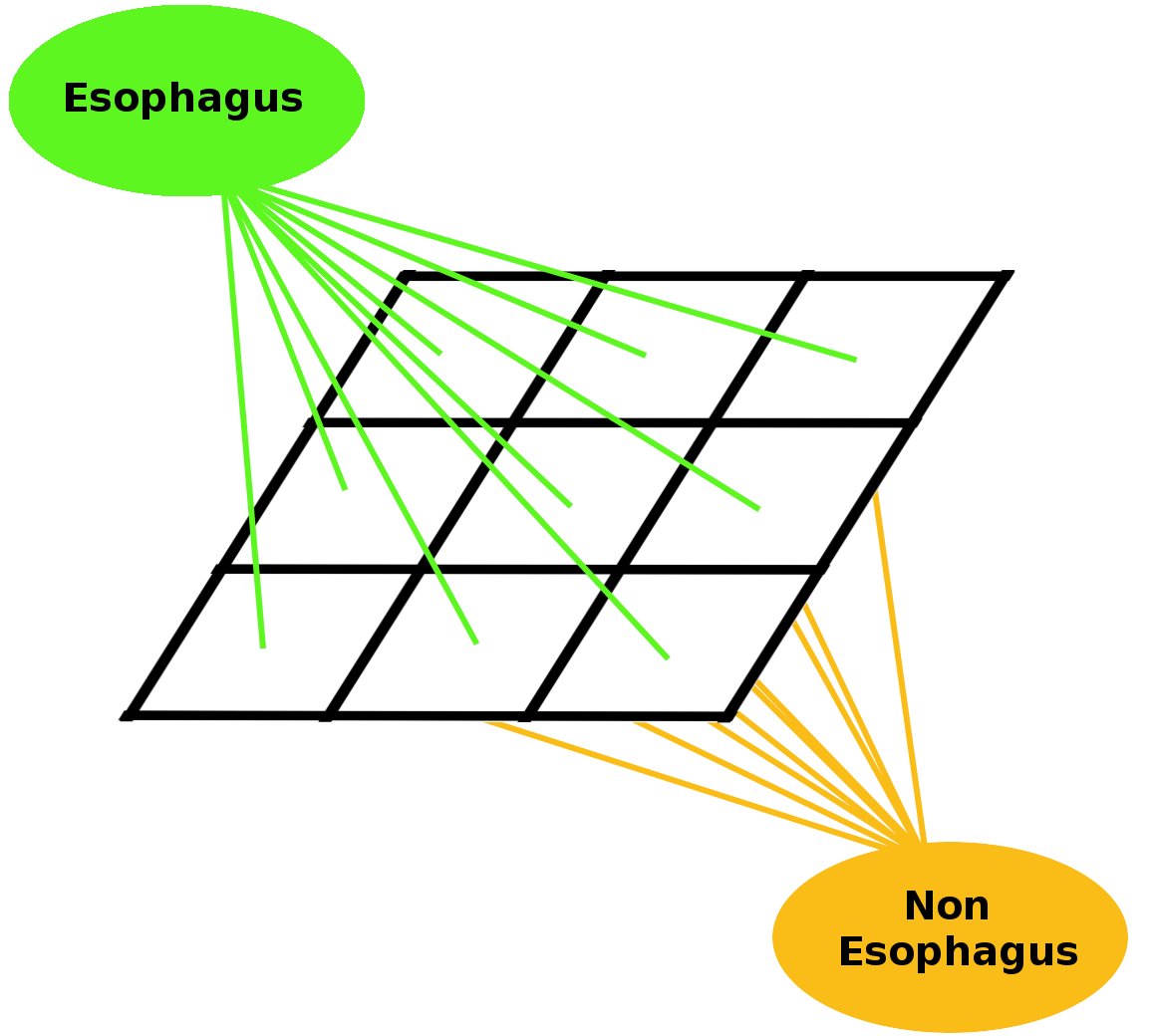} \\
        b) generated RW graph
        }
      
        }
        \caption{In a) the architecture of the proposed algorithms can be seen, b) shows the graph generated for the RW.}
\label{fig:architecture}
\end{center}        
\end{figure}

The proposed segmentation approach is three fold. First, a probability map of the esophagus is generated with a 3D fully CNN. Then, an active contour model (ACM) is fitted on the probability map for a first estimation of the esophageal centerline. Finally, probability map and centerline are used in combination with Hounsfield Unit and image gradient information to drive the RW towards the final contour. The architecture of our segmenation approach is depicted in Fig. \ref{fig:architecture} a. For evaluation, we run extensive experiments on a publicly available dataset as well as on clinical CTs used in RTP.

\subsection{Prior semantic esophagus segmentation}

Inspired by the outstanding performance of convolutional neural networks in medical image analysis, we employed the same architecture than in \cite{dolz20163d}, but including the dual re-scaled pathway shown in \cite{kamnitsas2016efficient}. Thus, our CNN is composed of 13 layer levels in total: 9 convolutional layers in each path, 3 fully-connected layers, and the classification layer. The number of kernels in each convolutional layer, from shallow to deeper, is as follows:  25, 25, 25, 50, 50, 50, 75, 75 and 75, sizes of which are equal to \vold{3}. As in standard CNNs, fully-connected layers are added at the end of the network to encode semantic information. Particularly for this work, three fully-connected layers, composed of 400, 200 and 150 hidden units were added. To ensure that the network contains only convolutional layers, we use the strategy described in \cite{kamnitsas2016efficient} and \cite{dolz20163d}, in which fully-connected layers are converted to a large set of \vold{1} convolutions. By doing this the network can retain spatial information and learn the parameters of these layers as in other convolutional layers. A down-sampling factor of 2 along the three dimensions is employed in the second path of the architecture (see \cite{kamnitsas2016efficient} for a detailed explanation of the dual-path approach followed) to capture a larger receptive field, which leads a broader view of the context. In the dual path both number and size of kernels is the same than in the main path. Due to computation and memory limitations, our network can not apply dense training over the whole 3D input volume. Instead, volumes are split into $S$ smaller sub-volumes, allowing dense inference on these samples.

A stride may also be defined for each convolutional layer, representing the displacement of the filter, along the three dimensions, after each application. To preserve spatial resolution, we use a unit stride for all convolutional layers. Each convolutional layer is followed by a Parametric Rectified Linear Unit (PReLU) \cite{he2015delving}, which applies an element-wise activation function. Let's define $\theta$ as the network trainable parameters (i.e., convolution filter weights, biases and PReLU activations in our case), and denote as $\mathcal{L}$ the set of ground-truth labels such that $L^v_s \in \mathcal{L}$ represents the label of voxel $v$ in the $s$-th sampled sub-volume. Following \cite{dolz20163d}, we defined the cost function as
\begin{equation}
\label{eq:cost}
\qquad \qquad  J(\theta; \mathcal{L}) \ = \ 
    -\frac{1}{S\!\cdot\!V} \sum^{S}_{s=1} \sum^{V}_{v=1} \log \, p_{L^v_s}(X_v),
\end{equation}
where $p_c(X_v)$ is the output of the classification layer for voxel $v$ and class $c$ (i.e. softmax output). As in \cite{dolz20163d} we set the sampled sub-volume size in our network to \vold{27} for training and \vold{45} for testing.

The optimization of network parameters is performed via RMSprop optimizer \cite{tieleman2012lecture}, using the cost function defined in (\ref{eq:cost}). Momentum was set to 0.6 and the initial learning rate to 0.001, being reduced by a factor of 2 after every 5 epochs (starting from epoch 10). In our work, weights in layer $l$ were initialized based on a zero-mean Gaussian distribution of standard deviation  $\sqrt{2/n_l}$, where $n_l$ denotes the number of connections to units in that layer. Our 3D FCNN was trained for 25 epochs, each one composed of 20 subepochs. At each subepoch, a total of 500 samples were randomly selected from the training images, and processed in batches of size 5. Our CNN is based on the work of \cite{dolz20163d} and \cite{kamnitsas2016efficient}, which architecture was developed using Theano \cite{bergstra2010theano}. We employed a server equipped with a NVIDIA Tesla P100 GPU with 16 GB of memory. Training took approximately 30 min per epoch, and around 13 hours for the fully trained CNN.

\subsection{Active Contour Model}

Even though the CNN generates a satisfactory estimation of the esophagus position, its result still contains several false positive areas with high probabilities and false negative areas with low probabilities. To prevent those areas from misleading the RW, an active contour model \cite{Kass1988} (ACM) was fitted on the CNN probability map. The model consists of several points, each placed on one slice of the CNN probability map. During optimization the points are attracted by high probability areas but restricted in their movement by a smoothness term. The result of the fitting process is a smooth continuous line from the first to the last slice connecting the high probability areas of the CNN probability map. The line could be seen as an expectation of the esophageal centerline.

\begin{figure}[t!]
     \begin{center}
     \mbox{
      \shortstack{
        \includegraphics[width=0.24\linewidth,height=0.24\linewidth]{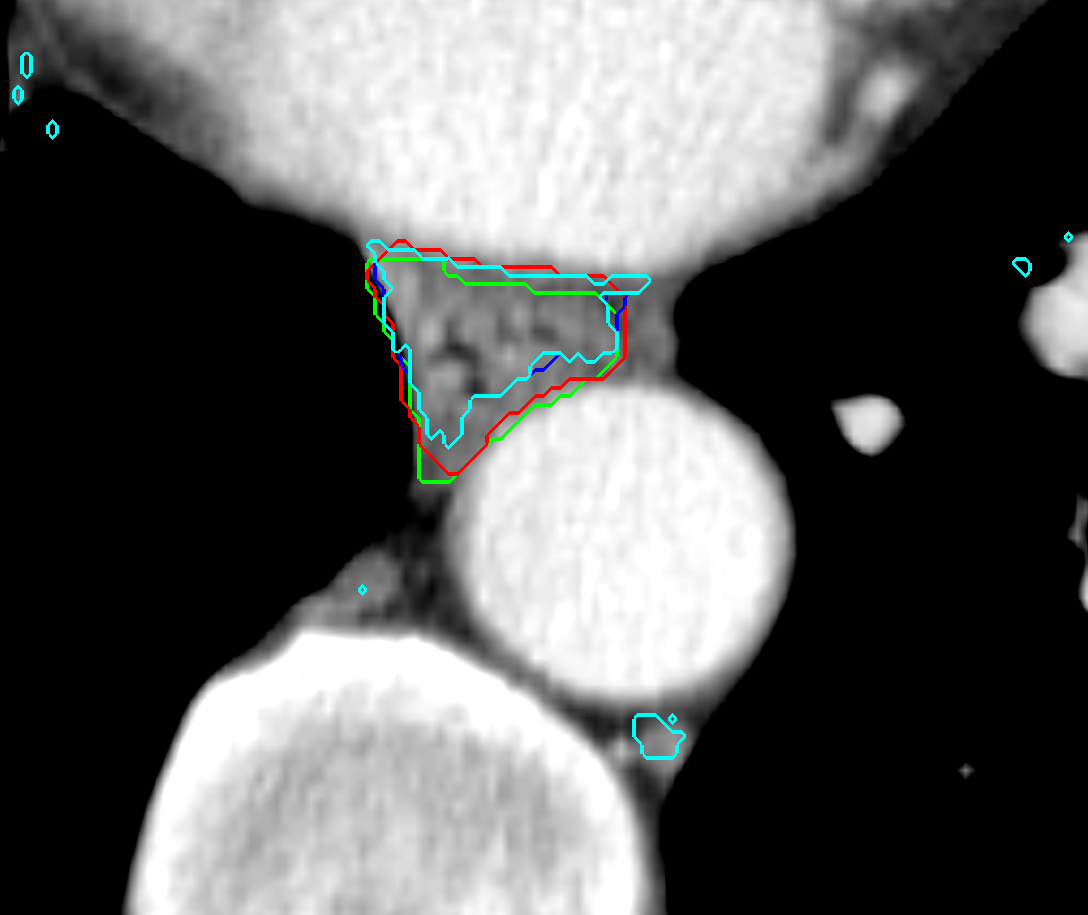} \\
        a) CT
        }
        \hspace{-1.5 mm}
         
      \shortstack{     
        \includegraphics[width=0.24\linewidth,height=0.24\linewidth]{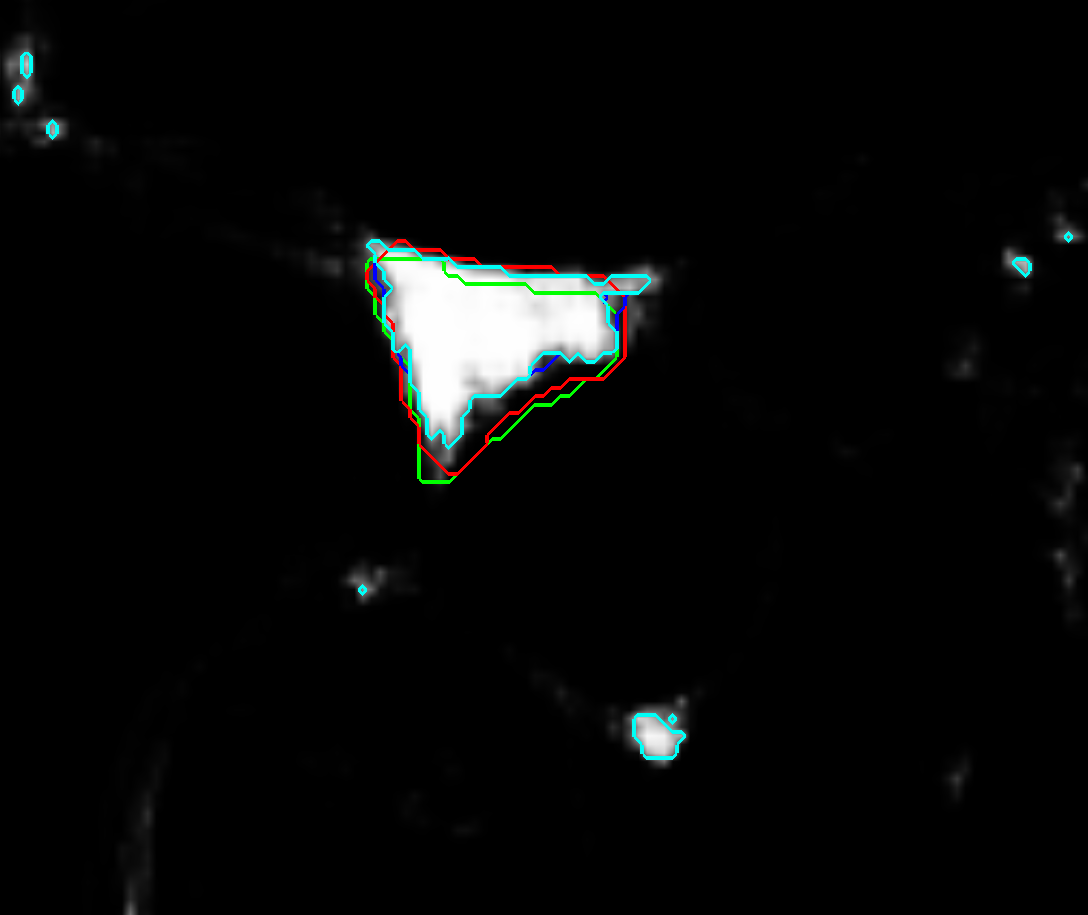} \\
        b) CNN prior}
        \hspace{-1.5 mm}

        \shortstack{
        \includegraphics[width=0.24\linewidth,height=0.24\linewidth]{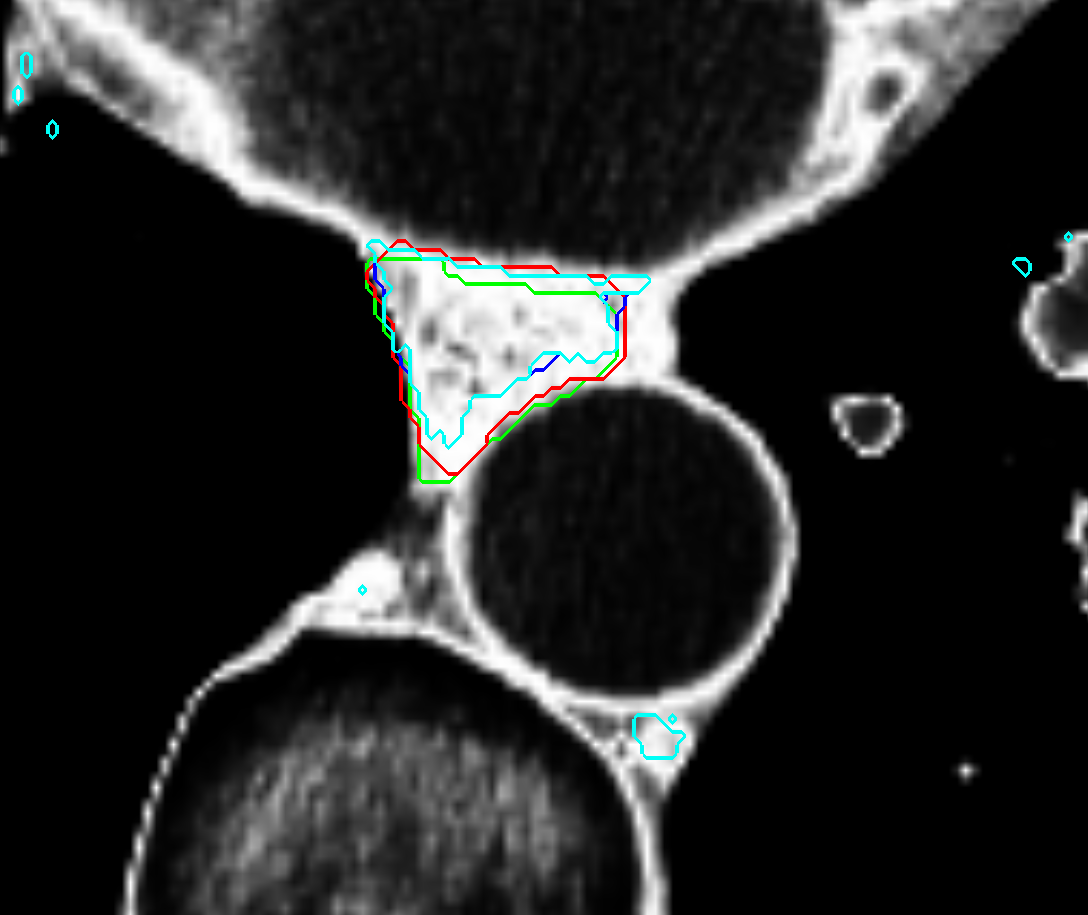} \\
        c) Intensity based prior}
        \hspace{-1.5 mm}
        
        \shortstack{
        \includegraphics[width=0.24\linewidth,height=0.24\linewidth]{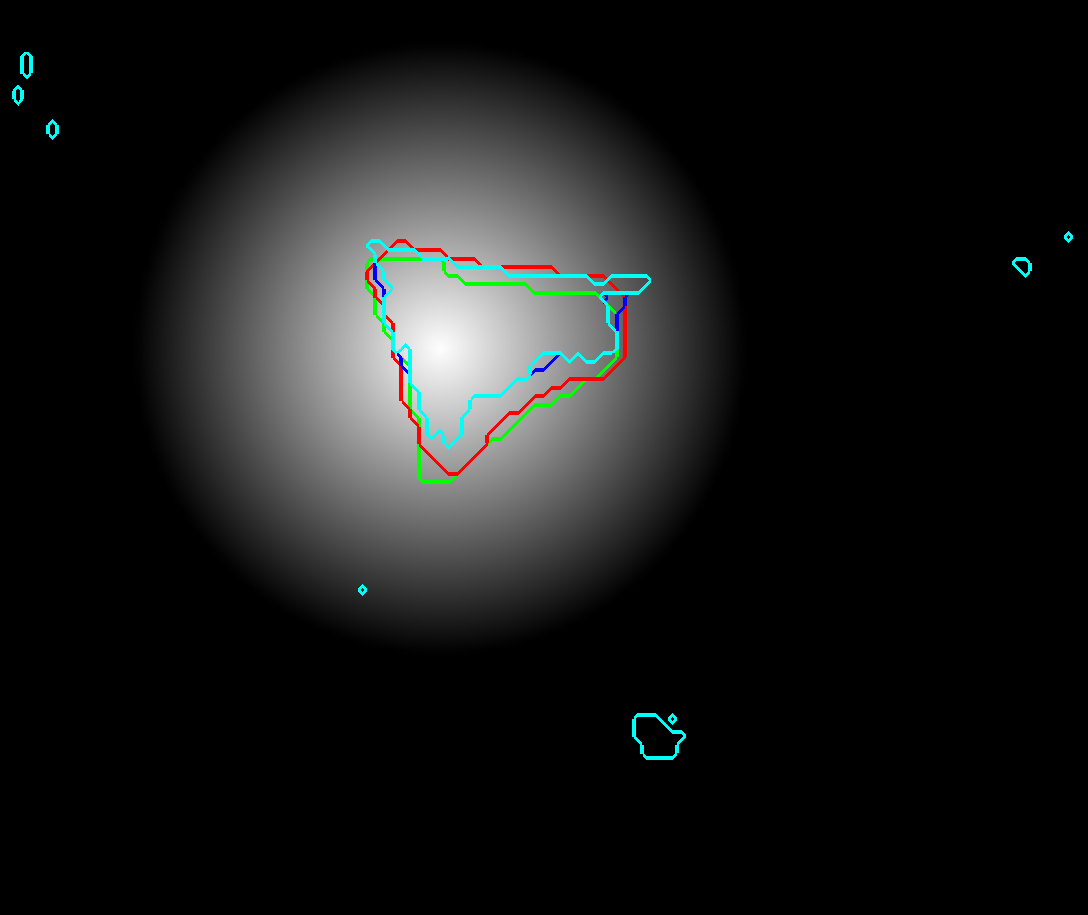} \\
        d) ACM distance map}

        }
        
       \mbox{
      \shortstack{
        \includegraphics[width=0.24\linewidth,height=0.24\linewidth]{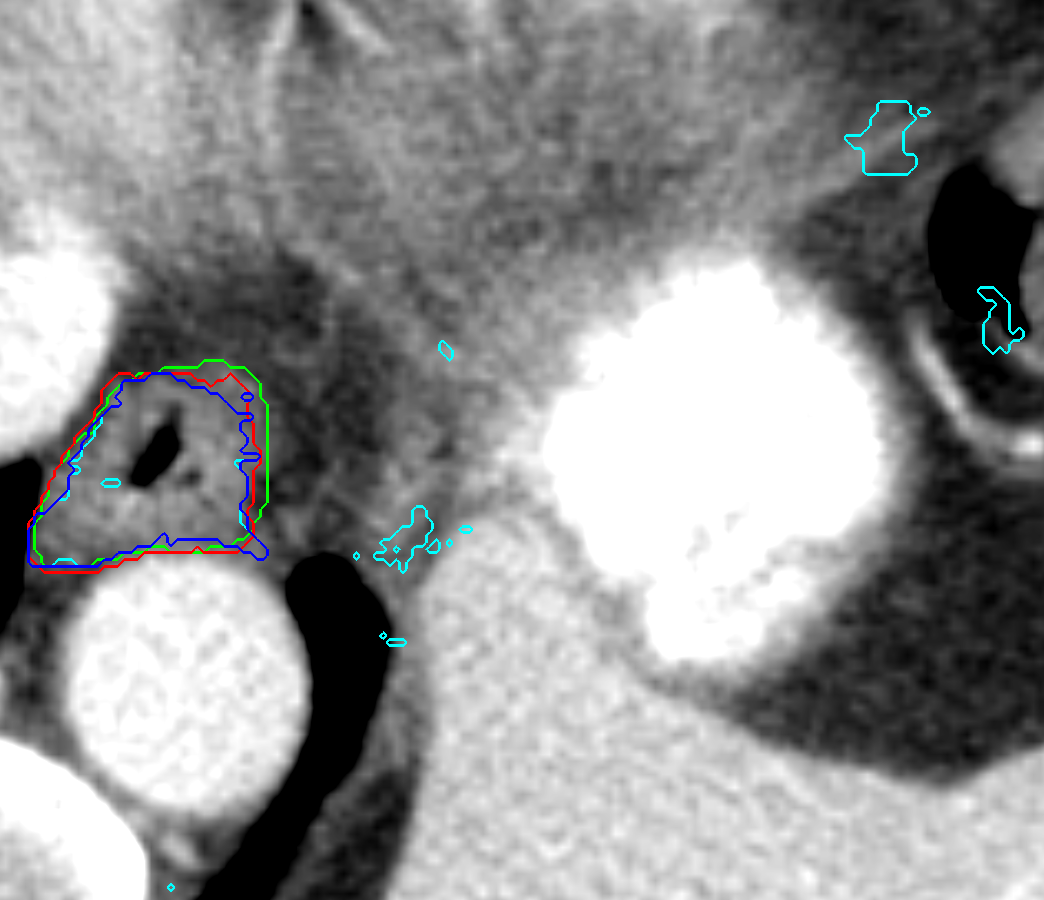} \\
        e) CT
        }
        \hspace{-1.5 mm}
         
      \shortstack{     
        \includegraphics[width=0.24\linewidth,height=0.24\linewidth]{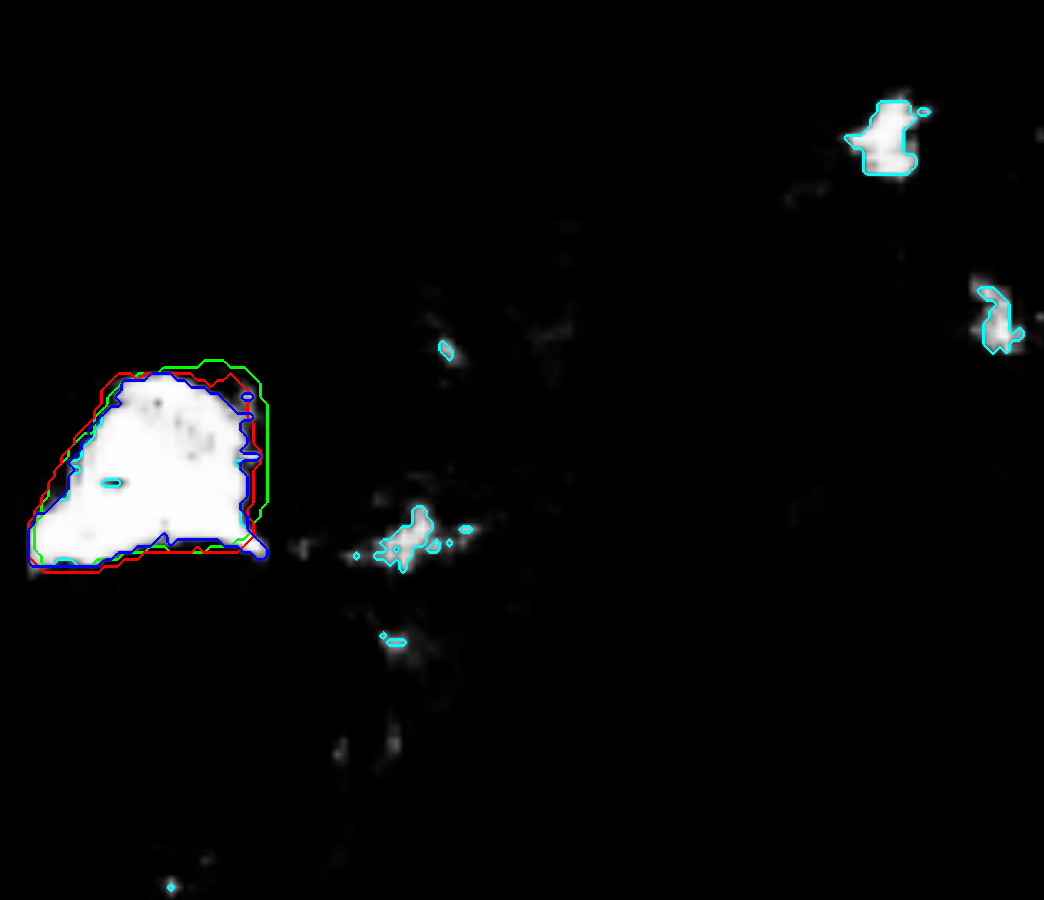} \\
        f) CNN prior}
        \hspace{-1.5 mm}

        \shortstack{
        \includegraphics[width=0.24\linewidth,height=0.24\linewidth]{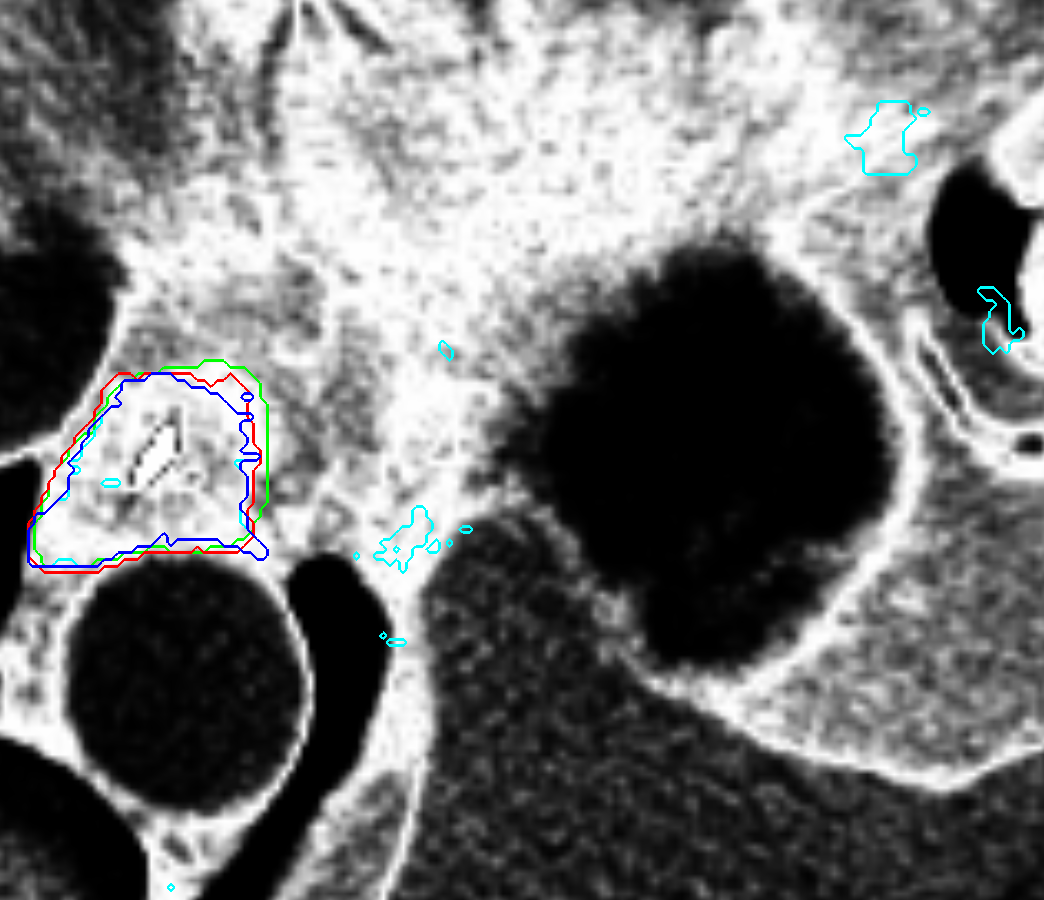} \\
        g) Intensity based prior}
        \hspace{-1.5 mm}
        
        \shortstack{
        \includegraphics[width=0.24\linewidth,height=0.24\linewidth]{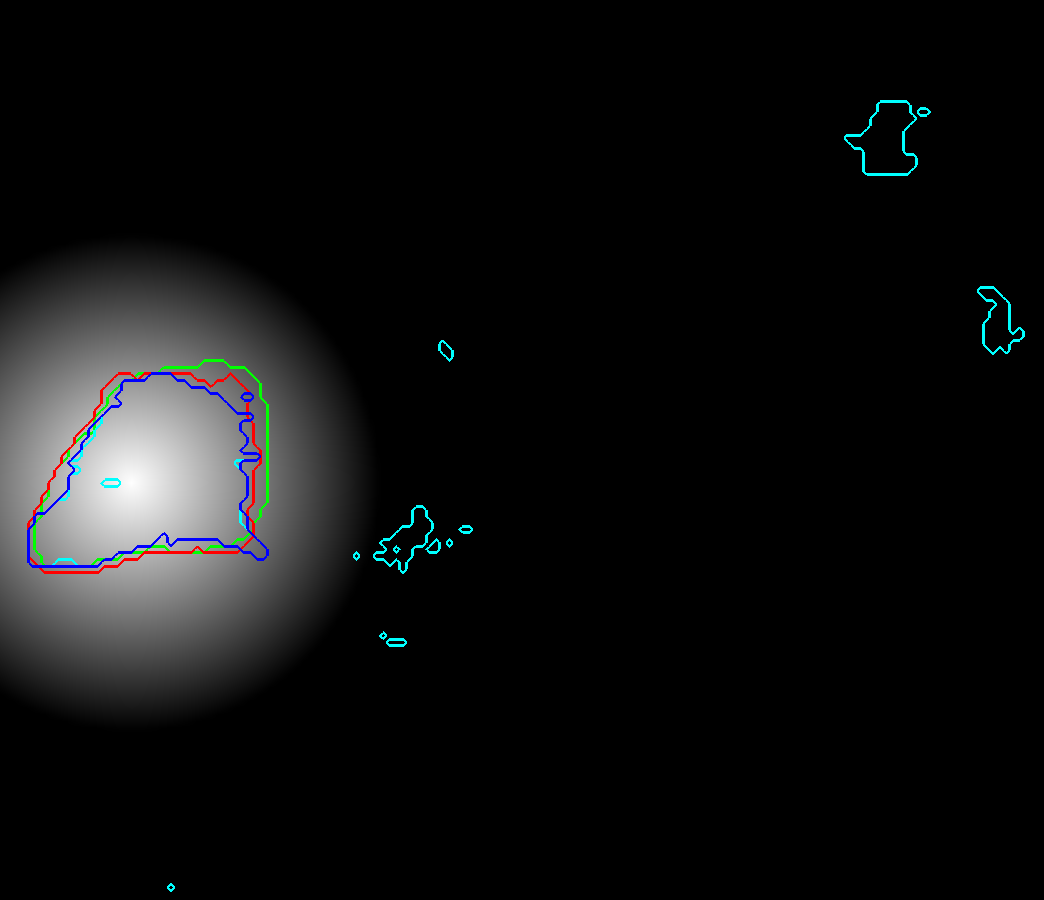} \\
        h) ACM distance map}
}
        
        \mbox{
      \shortstack{
        \includegraphics[width=0.24\linewidth,height=0.24\linewidth]{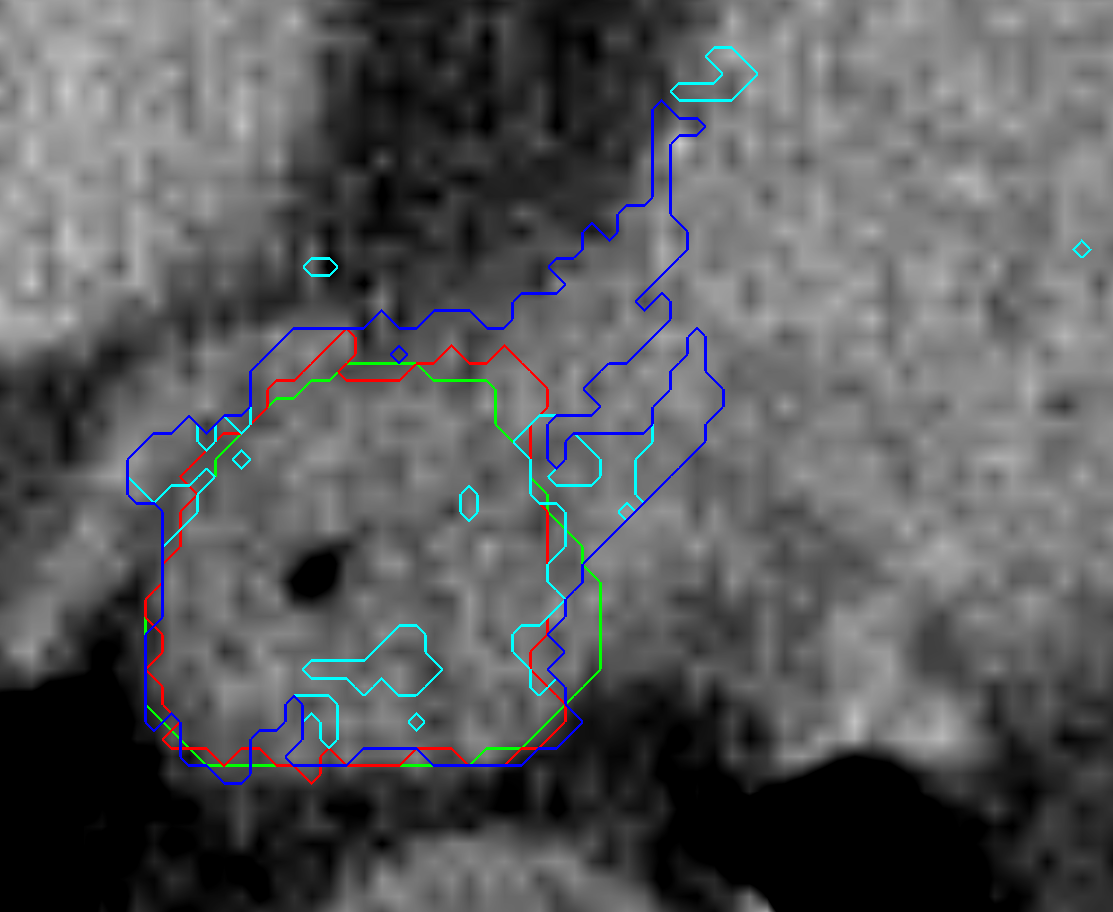} \\
        i) CT
        }
        \hspace{-1.5 mm}
         
      \shortstack{     
        \includegraphics[width=0.24\linewidth,height=0.24\linewidth]{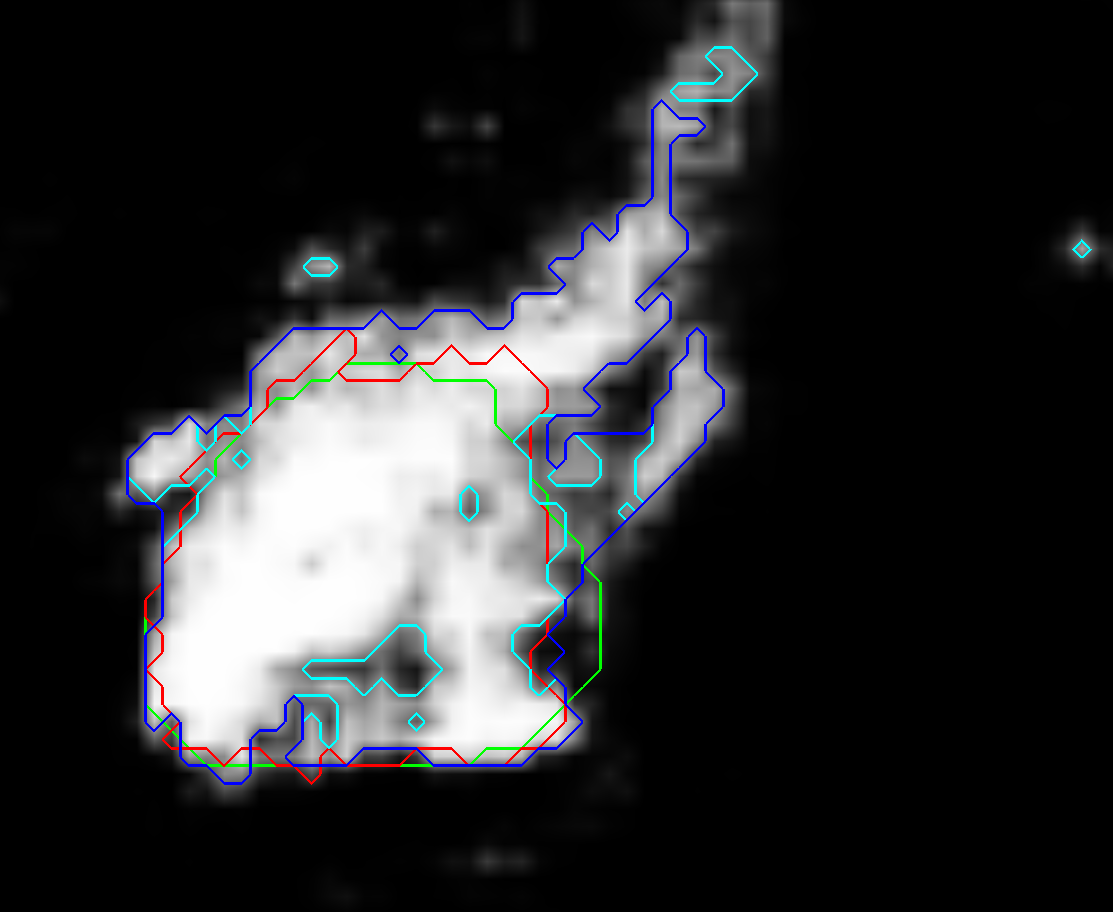} \\
        j) CNN prior}
        \hspace{-1.5 mm}

        \shortstack{
        \includegraphics[width=0.24\linewidth,height=0.24\linewidth]{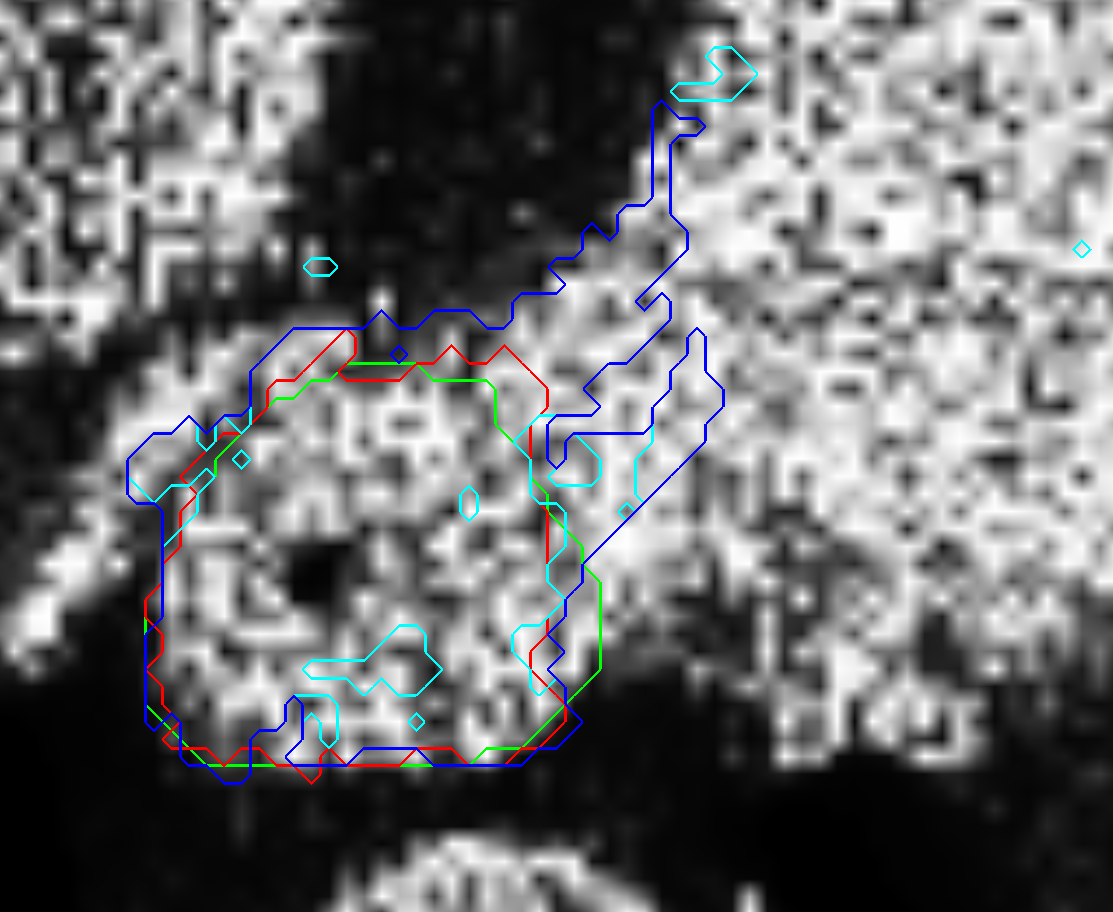} \\
        k) Intensity based prior}
        \hspace{-1.5 mm}
        
        \shortstack{
        \includegraphics[width=0.24\linewidth,height=0.24\linewidth]{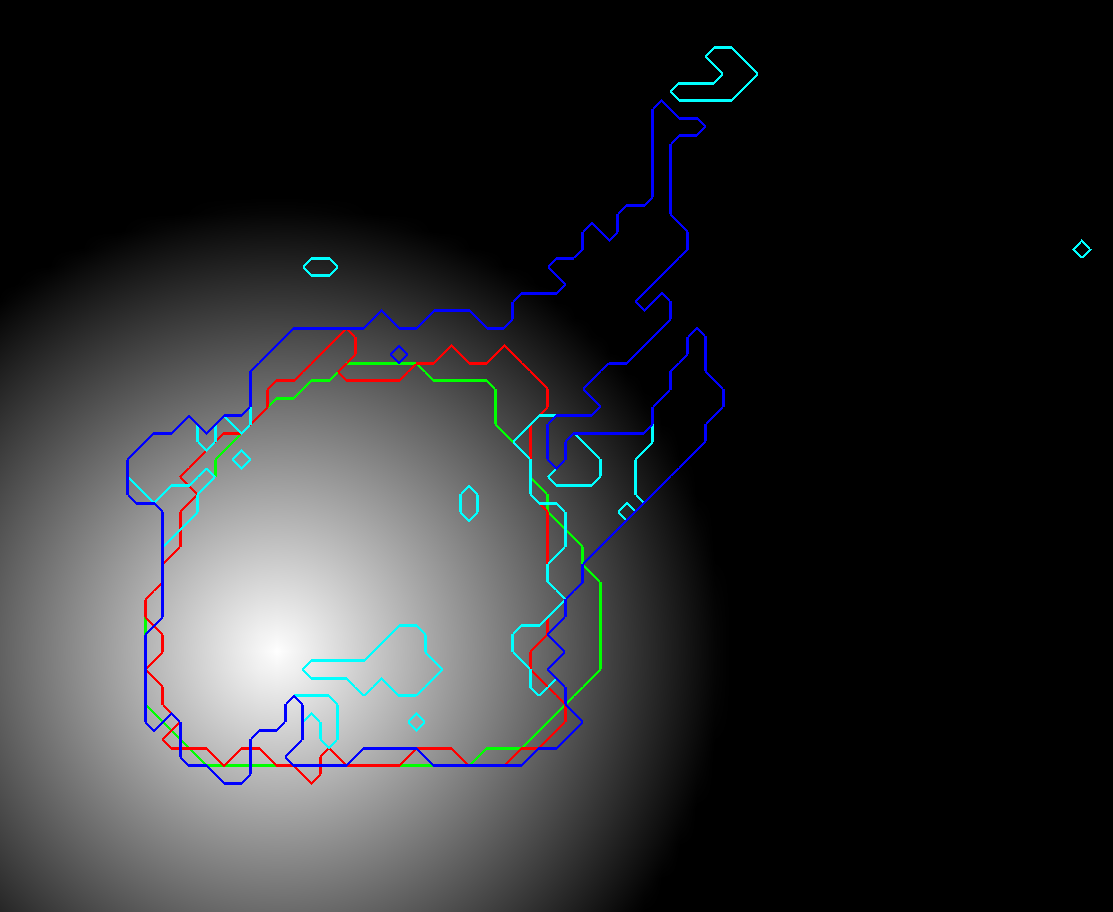} \\
        l) ACM distance map}

        }
        \caption{RW input images with reference contour (green), RW contour (red), CNN with 50 \% threshold (cyan) and CNN with 50 \% threshold and morphological closing (blue): first column depicts CT slices; second and third column show the probabilities generated by the CNN and the intensity prior model, respectively; the distance maps generated with the ACM are shown in the fourth colulmn. Figures a - d) show  public case 27, e - f) depict public case 01 and i - l) show clinical case 20.}
\label{fig:rwInput}
\end{center}        
\end{figure}

To make the centerline processable for the RW algorithm, it is transformed to a distance map with a value of 1 at the centerline and linearly decreasing values, which reach a minimum of 0 at a distance of 25 mm from the centerline. An example can be seen in Fig. \ref{fig:rwInput}d. The radius of 25 mm was chosen as it generously covers the usual width of the esophagus (usually 15 to 20 mm \cite{pmid21477778}) but excludes other unwanted structures.

\vspace{-3mm}
\subsection{Random Walk}

To generate a final contour on basis of the CNN generated probability map, the ACM distance map and the CT image data a random walk algorithm (RW) incorporating prior models \cite{gradyML} was chosen. This algorithm was selected because it is able to join information from non spatial data like intensity models and spatial correlations like voxel neighborhoods, and in contrast to the original work \cite{gradyRW} needs no input labels. The algorithm constructs a weighted undirected graph. In the graph each voxel is represented by a vertex and connected to its direct adjacent neighbors by a weighted edge. Additionally, for every segmentation class (esophagus and non esophagus) a vertex is added to the graph and connected to all voxel vertices (see Fig. \ref{fig:architecture} b). With the use of the Laplace matrix of the graph, a system of linear equations can be created and solved to calculate the probabilities for all voxels to belong to a given class. A detailed mathematical derivation can be found in \cite{gradyML} and \cite{gradyRW}.

The proposed RW makes use of CT image gradients and a prior model. Edge weights between adjacent voxel vertices are calculated with the use of the CT image gradients by:
\begin{equation}
w(e_{ij}) = \frac{\frac{1}{{\sigma_{\delta} \sqrt {2\pi } }}e^{{{ - \left( {\delta_{ij} - \mu_{\delta} } \right)^2 } \mathord{\left/ {\vphantom {{ - \left( {x - \mu_{\delta} } \right)^2 } {2\sigma_{\delta} ^2 }}} \right. \kern-\nulldelimiterspace} {2\sigma_{\delta} ^2 }}}}{Z},
\end{equation}
where $\delta_{ij}$ is the intensity difference between voxels represented by vertex $i$ and $j$, $\sigma_{\delta}$ and $\mu_{\delta}$ are standard deviation and mean of the gradients inside the esophagus contours of the learning datasets. $Z$ is a normalization constant to scale the highest value of $w(e_{ij})$ to 1 and the smallest value to 0. 
The prior probability model used incorporates information from the CNN, the ACM and the CT image intensity distribution. For this, two vertices, representing the the two segmentation classes,  are added to the graph and connected to all voxel vertices by edges. The weights for the esophagus class and non esophagus class are calculated by: 
\begin{equation}
w_{esophagus}(e_{i}) = \prod_j p_{j,i}
\label{eq:esophagusProb}
\end{equation}
and
\begin{equation}
w_{non\, esophagus}(e_{i}) = \prod_j 1 - p_{j,i}
\label{eq:nonesophagusProb}
\end{equation}
with $j \in \lbrace CNN, ACM, CT \rbrace$. The factors $p_{CNN,i}$ and $p_{ACM,i}$ are the values of voxel $i$ in the CNN probability map or the ACM distance map, respectively.

The third factor of the prior model reflects the probability of a CT voxel with a given Hounsfield Unit value being part of the esophagus and is calculated with the use of a Gaussian mixture model that was fitted to the Hounsfield Unit distribution of the reference contours in the training data. 

The RW and the ACM algorithms base both on the respective python scikit-image library\cite{scikit-image} implementation. The scikit implementations were modified or extended to meet our requirements.

\subsubsection{Pre- and Postprocessing}
In a preprocessing step all CT voxels with HU lower than -150 were set to the mean esophageal voxel intensity of the training data to eliminate high gradients in the esophagus due to enclosed air and facilitate the inclusion of this region by the RW. The final contour was generated by applying a morphological closing operator to eliminate false positive spots on the RW result.

\subsection{Datasets and Analysis}

Evaluation of the proposed algorithm was done on 20 clinical and 30 publicly available datasets. All datasets consisted of a CT and a reference contour. The results of the proposed algorithm were compared to the reference contour with regard to spatial overlap and shape. Additionally, statistical analysis with the Wilcoxon signed rank test was performed to evaluate the impact of the RW step on the result. Null hypothesis was a median difference of zero between the compared groups and alpha was set to 1 \%. The Wilcoxon test was chosen due to non normal distribution and heteroscedasticity of the data.

\subsubsection{Datasets}

The 20 clinical subjects were acquired on a Philips GEMINI TF Big Bore or on a Siemens Emotion with voxel resolutions from \nobreak{$0.78 \times 0.78 mm$} to \nobreak{$1.17 \times 1.17 mm$} in axial plane and $2.00 \text{ to } 3.00 mm$ in z-direction. The image matrix had a dimension of \nobreak{$512 \times 512$} in x- and y-direction and $60$ to $181$ slices. All subjects  depicted  the  whole esophagus from the upper sphincter (cricoid level) down to the gastro-esophageal junction. Image acquisition was performed while the patient was freely breathing. The  reference  contour  was created  by one experienced  radiation oncologist according to the EORTC 22113-08113 Lungtech protocol \cite{adebahr2015lungtech} and peer reviewed by at least one other expert. Additionally, 30 publicly available subjects from the ''Multi-Atlas Labeling Beyond the Cranial Vault - Workshop and Challenge'' \footnote{https://www.synapse.org/$\#$!Synapse:syn3193805/wiki/89480} were used. The CT resolution ranged from \nobreak{$0.67 \times 0.67 mm$} to \nobreak{$0.98 \times 0.98 mm$} in x- and y-direction and $3.00 \text{ to } 5.00 mm$ in z-direction. The image matrix had a dimension of \nobreak{$512 \times 512$} in x- and y-direction and $50$ to $198$ slices. On all subjects only the lower half of the esophagus is depicted. The reference contours are available within this dataset.

\subsubsection{Evaluation}

Different proposed methods have lacked a consensus on evaluation in terms of comparison metrics. Since each metric yields different information, their choice is important and must be considered in the appropriate context. The S\o{}rensen--Dice index (DSC) \cite{sorensen1948method} has been widely employed to compare volume similarities. The DSC for a volume A and a volume B is calculated by: 
\begin{equation}
DSC = \dfrac{2\left(A \cap B \right)}{A+B}
\end{equation}

Nevertheless, volume-based metrics are very insensitive to edge differences when they have a small impact on the overall volume. This makes that two segmentations showing a high degree of spatial overlapping might present clinically relevant differences at the edges. This is particularly important in RTP, where the contours serve as critical input to compute the delivered dose. An analysis on shape fidelity of the segmentation outline is therefore highly recommended, since any underinclusion on the OAR segmentation might lead to a part of the healthy tissue exposed to radiation. Thus, distance-based metrics -Hausdorff distance (HD) and average symmetric surface distance (ASSD) - were also employed. 





\section{Results}



Evaluation was done in three steps. First, the benefit of the RW step was inspected. For this the RW contour was compared to the 50 \% threshold contour of the CNN probability map and a morphologically closed version of the 50 \% contour to exclude small false positive spots. Second, the esophagus contours as a whole were analyzed. In a third step, we cut the datasets below the lower tip of the heart and analyzed only the upper part of the organ. This was done because for the best performing fully automatic algorithm so far published \cite{feulner2011probabilistic} performance was analyzed only in the region from the thyroid gland down to a level below the left atrium and from our perspective the lower part of the esophagus until the gastro-esophageal junction is  a very challenging one to delineate, as the esophagus changes its shape in this region (compared to the upper part, the diameter gets larger) and the transfer from esophagus to stomach is hardly visible.

We want to emphasize that a first analysis of the results showed up that probability values generated by the CNN for patient 7 in synapse dataset were close to zero in nearly the whole image. A deeper inspection revealed that this case is the only one presenting intense stripe shaped artifacts (an example is depicted in Fig. \ref{fig:discussion}a). As these artifacts are only visible in one subject, the CNN was not able to learn how to deal with them. Therefore, this subject was excluded in further evaluations.

In table \ref{table:mainTable}, the mean values for the S\o{}rensen–Dice index (DSC), average symmetric surface distance (ASSD) and Hausdorff distance (HD) between reference contours and the CNN generated esophagus outlines (with and without morphological closing) as well as the RW processed CNN contours for both datasets are depicted. Results in the whole volume, as well as in the region ranging from the upper sphincter down to the gastro-esophageal junction are presented.

\begin{table}[]
\begin{scriptsize}
\centering
\begin{tabular}{lccccccccc}
\hhline{==========}
\multicolumn{10}{c}{\textbf{Synapse Dataset}}                                           \\
  & \multicolumn{3}{c}{\textbf{CNN}} 
  & \multicolumn{3}{c}{\textbf{CNN + Closing}}  
  & \multicolumn{3}{c}{\textbf{Proposed}}                           
  \\
  & \multicolumn{1}{l}{\textbf{DSC}} & \multicolumn{1}{l}{\textbf{ASSD (mm)}} & \multicolumn{1}{l}{\textbf{HD (mm)}} & \multicolumn{1}{l}{\textbf{DSC}} & \multicolumn{1}{l}{\textbf{ASSD (mm)}} & \multicolumn{1}{l}{\textbf{HD (mm)}} & \multicolumn{1}{l}{\textbf{DSC}} & \multicolumn{1}{l}{\textbf{ASSD (mm)}} & \multicolumn{1}{l}{\textbf{HD (mm)}} \\
  \hline
\multicolumn{10}{c}{Whole Volume}                                                   
\\
\textbf{mean} & 0.53  & 21.39  & 177.66  & 0.62 & 10.48 & 76.82 & 0.73 & 1.77 & 14.42   \\
\textbf{std}  & 0.15  & 9.65   & 29.10  & 0.16   & 12.15   & 64.65  & 0.12   & 1.17  & 7.79   \\
\multicolumn{10}{c}{Cropped Volume}   \\
\textbf{mean} & 0.55  & 22.23   & 177.21  & 0.66  & 10.40   & 71.93   & \textbf{0.76}  & \textbf{1.29}  & \textbf{10.68} \\
\textbf{std}  & 0.17  & 11.00  & 28.75  & 0.18   & 13.51  & 67.22   & 0.13   & 1.03   & 6.11   \\
\multicolumn{10}{l}{}                                                                   \\
   \hline
\multicolumn{10}{c}{\textbf{Own clinical Dataset}}                                     \\
 & \multicolumn{3}{c}{\textbf{CNN}} 
 & \multicolumn{3}{c}{\textbf{CNN + Closing}}
 & \multicolumn{3}{c}{\textbf{Proposed}}                                               \\
 & \textbf{DSC}  & \textbf{ASSD (mm)}  & \textbf{HD (mm)}  & \textbf{DSC}  & \textbf{ASSD (mm)}  & \textbf{HD (mm)}  & \textbf{DSC}  & \textbf{ASSD (mm)}  & \textbf{HD (mm)}  \\
   \hline
\multicolumn{10}{c}{Whole Volume}                                                      \\
\textbf{mean} & 0.58  & 15.63  & 180.22 & 0.68  & 8.14   & 60.54  & 0.74  & 1.99  & 19.68  \\
\textbf{std}  & 0.16  & 9.17  & 23.96  & 0.15  & 13.71  & 63.88  & 0.10  & 1.34 & 12.32   \\
\multicolumn{10}{c}{Cropped Volume}                                                    \\
\textbf{mean} & 0.60  & 15.21  & 180.37  & 0.70  & 7.40  & 60.25   & \textbf{0.76}  & \textbf{1.47}  & \textbf{13.63}  \\
\textbf{std}  & 0.15  & 8.47  & 24.00  & 0.14  & 11.73  & 63.54  & 0.09  & 0.68  & 7.51 \\
\hhline{==========}
\end{tabular}
\caption{Mean values of DSC, ASSD and HD for the three configurations analyzed on both datasets. While first part of results represents segmentations on the whole volume, the second part deploys performance on cropped generated contours.}
\label{table:mainTable}
\end{scriptsize}
\end{table}

\begin{figure}[t!]
     \begin{center}
     \includegraphics[width=1\linewidth]{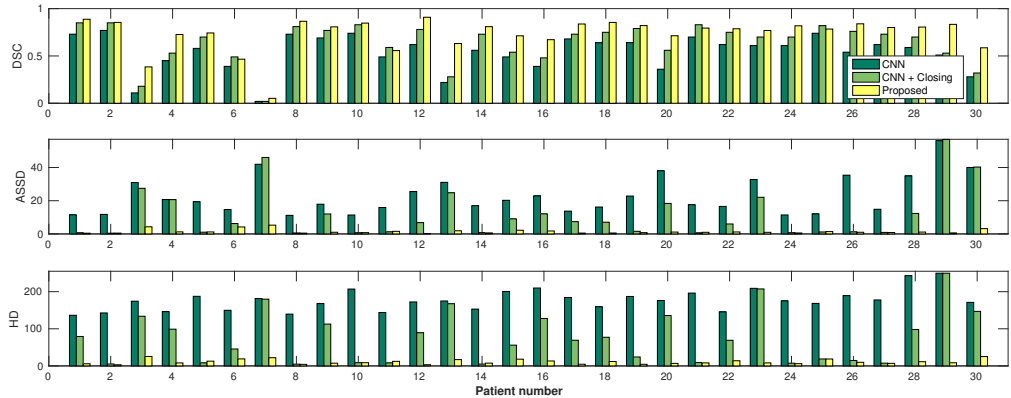}
     \caption{DSC, ASSD and HD values for the three pipelines across all the patients in Synapse dataset.}
     \label{fig:resSynapse}
     \end{center}
     
\end{figure}

\begin{figure}[t!]
     \begin{center}
     \includegraphics[width=1\linewidth]{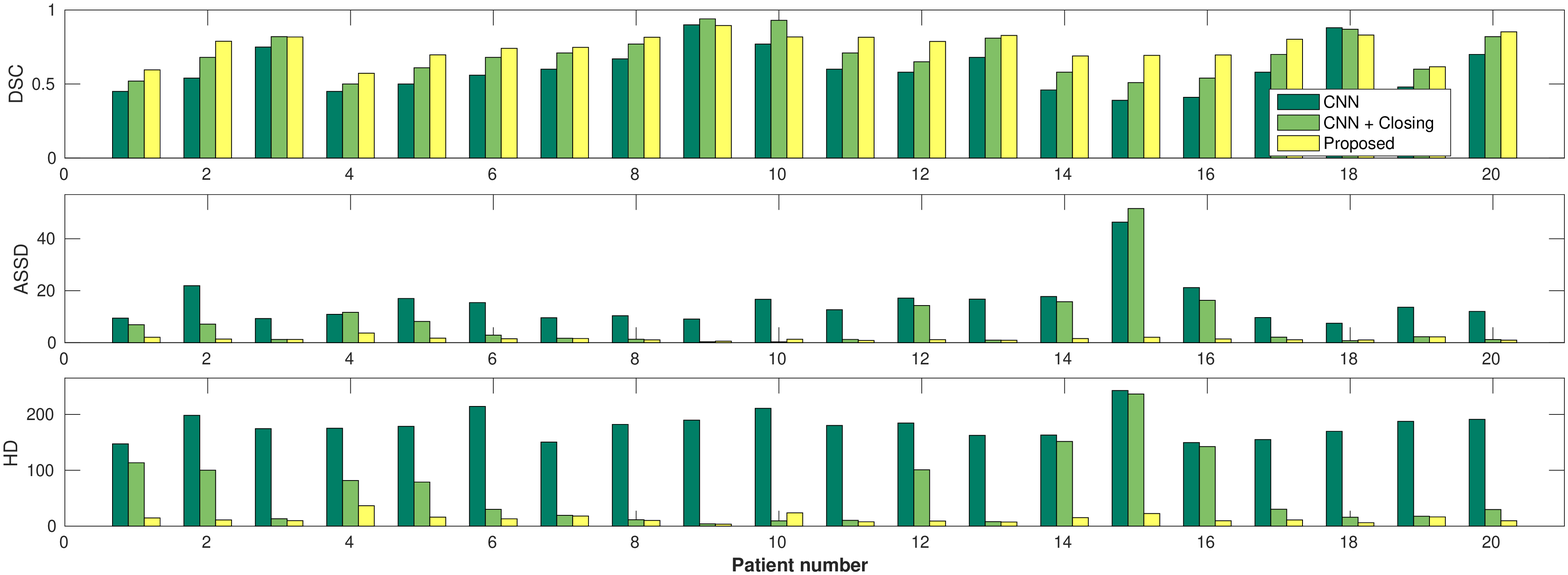}
     \caption{DSC, ASSD and HD values for the three pipelines across all the patients in the own clinical dataset.}
     \label{fig:resFreiburg}
     \end{center}
\end{figure}

From this table, it can be observed that driving the RW algorithm with CNN-based probability maps improves the final contours. This is supported by the detailed bar plots in figures \ref{fig:resSynapse} and \ref{fig:resFreiburg}, which depict the results for the three configurations across all the subjects in both datasets. An increase of DSC and a decrease of ASSD and HD were achieved by the RW algorithm for almost all cases. On average, an increase of 30$\%$ and nearly 10$\%$ on DSC were measured with respect to the baseline CNN, and the CNN + closing, respectively. On the other hand, decrease on shape-based metrics presented a higher value. For example, by guiding the RW with the probability maps, the ASSD was reduced between 80 and 90$\%$ with respect to the second configuration, i.e. CNN + closing. These findings were more noticeable in the cropped volumes. 

Statistical comparison of the results in table \ref{table:mainTable} is presented in table \ref{table:ANOVA}. It can be seen that the improvement of the contours by the RW is statistically significant for all figures but DSC and ASSD of the clinical datasets. This is due to some clinical cases (e.g. 9 and 10 in Fig. \ref{fig:resFreiburg}) where the RW worsened the result. When conducting statistical analysis over all datasets DSC, ASSD and HD improvements of the RW become highly significant. Therefore in the further analyses and comparisons only CNN + RW results are inspected. 
 
\begin{table}[]
\centering
\begin{tabular}{lcccccc}
\hhline{=======}
\multicolumn{7}{c}{\textbf{Statistical analysis results (p-values)}}                                    \\
                         & \multicolumn{2}{c}{DSC} & \multicolumn{2}{c}{ASSD} & \multicolumn{2}{c}{HD}  \\
                         & CNN     & CNN + Closing & CNN      & CNN + Closing & CNN     & CNN + Closing \\
\hline
\textbf{Synapse Dataset} & $<$0.01  & $<$0.01        & $<$0.01   & $<$0.01        & $<$0.01 & $<$0.01        \\
\textbf{Own Dataset}     & $<$0.01 & 0.14        & $<$0.01  & 0.08        & $<$0.01 & $<$0.01       \\
\hhline{=======}
\end{tabular}
\caption{Wilcoxon signed rank test comparing results generated by CNN + RW to CNN and  CNN + closing.}
\label{table:ANOVA}
\end{table}

For the cropped datasets the overall mean DSC was 0.76 $\pm$ 0.11. ASSD and HD had a mean of 1.36 $\pm$ 0.90 and 11.88 $\pm$ 6.80. For the whole esophagus values worsened slightly to a mean DSC of 0.73 $\pm$ 0.11, mean ASSD of 1.86 $\pm$ 1.23 and a mean HD of 16.57 $\pm$ 10.11.

Table \ref{table:Results} presents the results obtained by the proposed method in both datasets analyzed and puts our approach in context with other methods found in the literature. It can be seen that only one semi automatic approach \cite{rousson2006probabilistic} achieved better performance in terms of volume overlapping. Nevertheless, when considering distance based metrics our algorithm yielded the best results. For example, taking results from both datasets into account, the mean ASSD was reduced nearly 35$\%$ and 25$\%$ with respect to the best semi-automatic \cite{kurugol2011centerline} and the fully automatic approach \cite{feulner2011probabilistic}, respectively.

\begin{table}[ht!]
\begin{scriptsize}
\centering
\renewcommand{\arraystretch}{1.2}
\begin{tabular}{lccccc}
\hhline{======}
\toprule
\textbf{Method} & \textbf{FA/SA}** & \textbf{DSC} & \textbf{ASSD (mm)} & \textbf{HD (mm)} & \textbf{No. subjects} \\
\hline
 \midrule\midrule
\textbf{Proposed method (Own data)} & FA  & 0.76 $\pm$ 0.09  & 1.47 $\pm$ 0.68 & 13.63 $\pm$ 7.51  & 20  \\
\textbf{Proposed method (Synapse)}  & FA  & 0.76 $\pm$ 0.13  & 1.29 $\pm$ 1.03 & 10.68 $\pm$ 6.11 & 30  \\
(Feulner et al, 2011) \cite{feulner2011probabilistic}    & FA      & 0.74 $\pm$ 0.14  & 1.80 $\pm$ 1.17     & 12.62 $\pm$ 7.01         & 144          \\
(Rousson et al., 2006) \cite{rousson2006probabilistic}   & SA      & 0.80 $\pm$ --    & -- $\pm$ --         & -- $\pm$ --              & 20           \\
(Fieselmann et al., 2008) \cite{fieselmann2008esophagus} & SA      & 0.73 $\pm$ 0.14  & -- $\pm$ --         & -- $\pm$ --              & 8            \\
(Feulner et al., 2009) \cite{feulner2009fast}            & SA      & -- $\pm$ --  & 2.28 $\pm$ 1.58 & 14.5 $\pm$ --            & 117          \\
(Kurugol et al., 2011) \cite{kurugol2011centerline}      & SA      & -- $\pm$ --      & 2.10 $\pm$ 1.90     & 15.10 $\pm$ --           & 8            \\
(Grosgeorge et al., 2013) \cite{grosgeorge2013esophagus} & SA      & 0.61 $\pm$ 0.06  & -- $\pm$ --         & -- $\pm$ --              & 6            \\
\bottomrule
\multicolumn{6}{l}{**Fully Automated (FA) / Semi-Automated (SA)}\\
\hhline{======}
\end{tabular}
\caption{Comparison of esophagus segmentation performance between our proposed 3D FCNN-based approach and previous research. In the table S\o{}rensen–Dice index (DSC), Hausdorff distance (HD) and average symmetric surface distance (ASSD) are stated.}
\label{table:Results}
\end{scriptsize}
\end{table}

\begin{figure}[t!]
     \begin{center}
     \mbox{
      \shortstack{
        \includegraphics[width=0.3\linewidth]{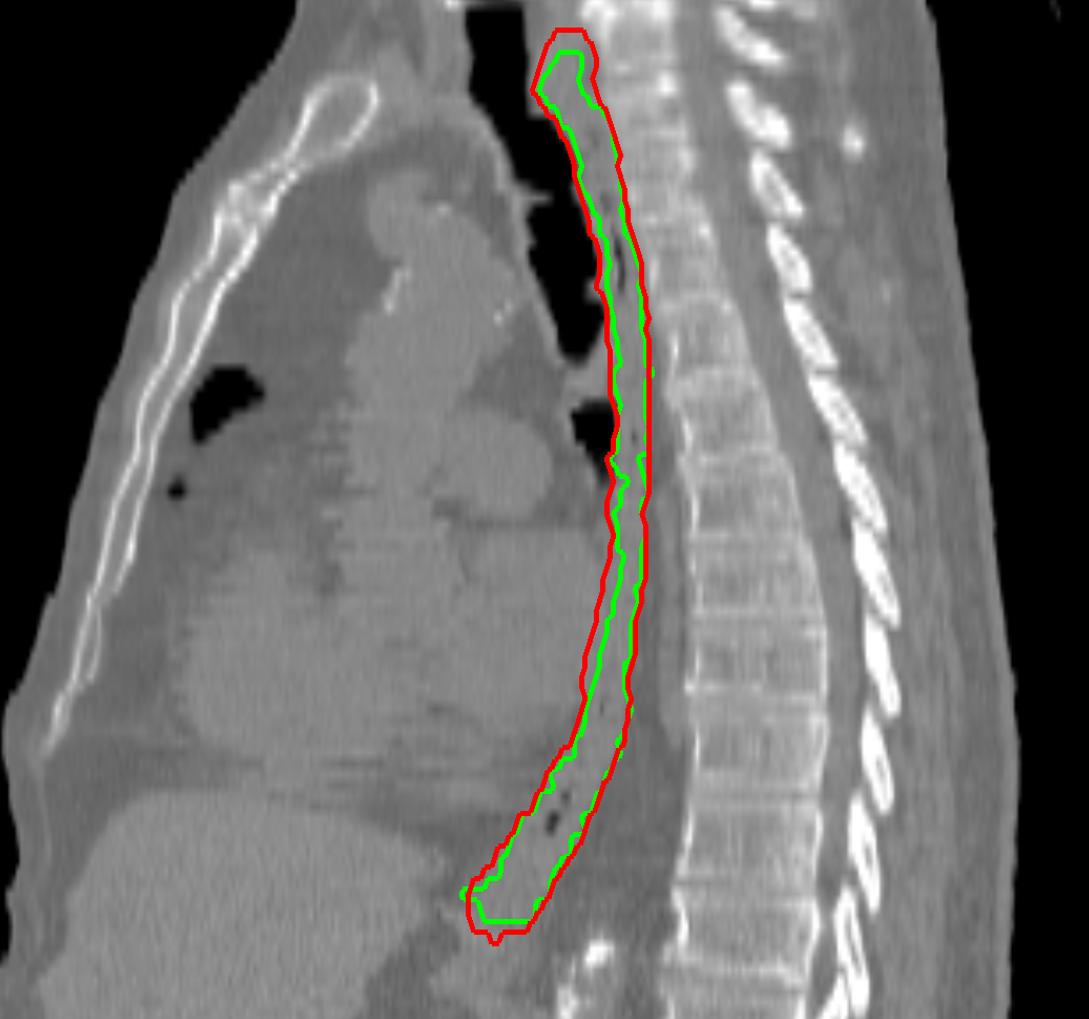} \\
        a) clinical case 09 \\ sagittal view
        }
        \hspace{-1.5 mm}
         
      \shortstack{     
        \includegraphics[width=0.3\linewidth]{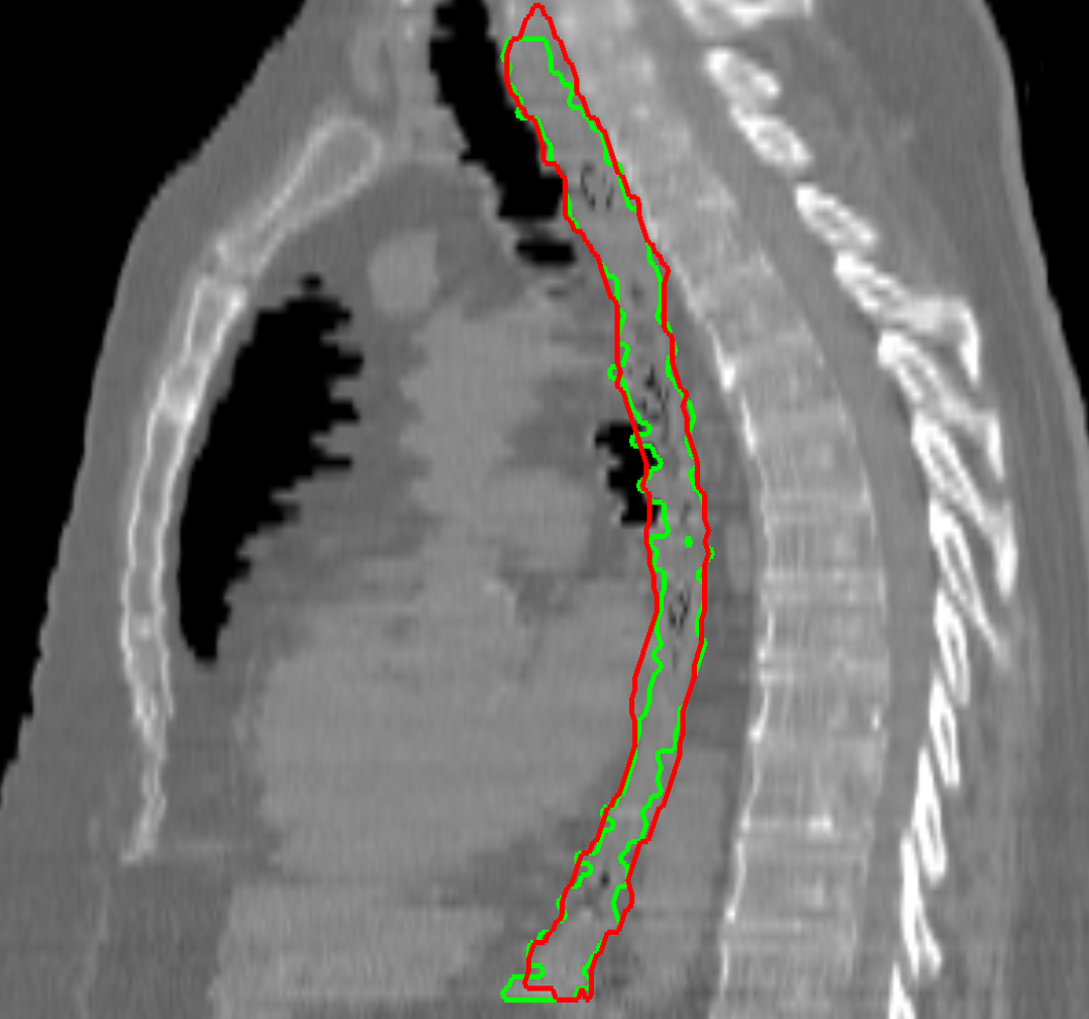} \\
        b) clinical case 13 \\ sagittal view}
        \hspace{-1.5 mm}

        \shortstack{
        \includegraphics[width=0.3\linewidth]{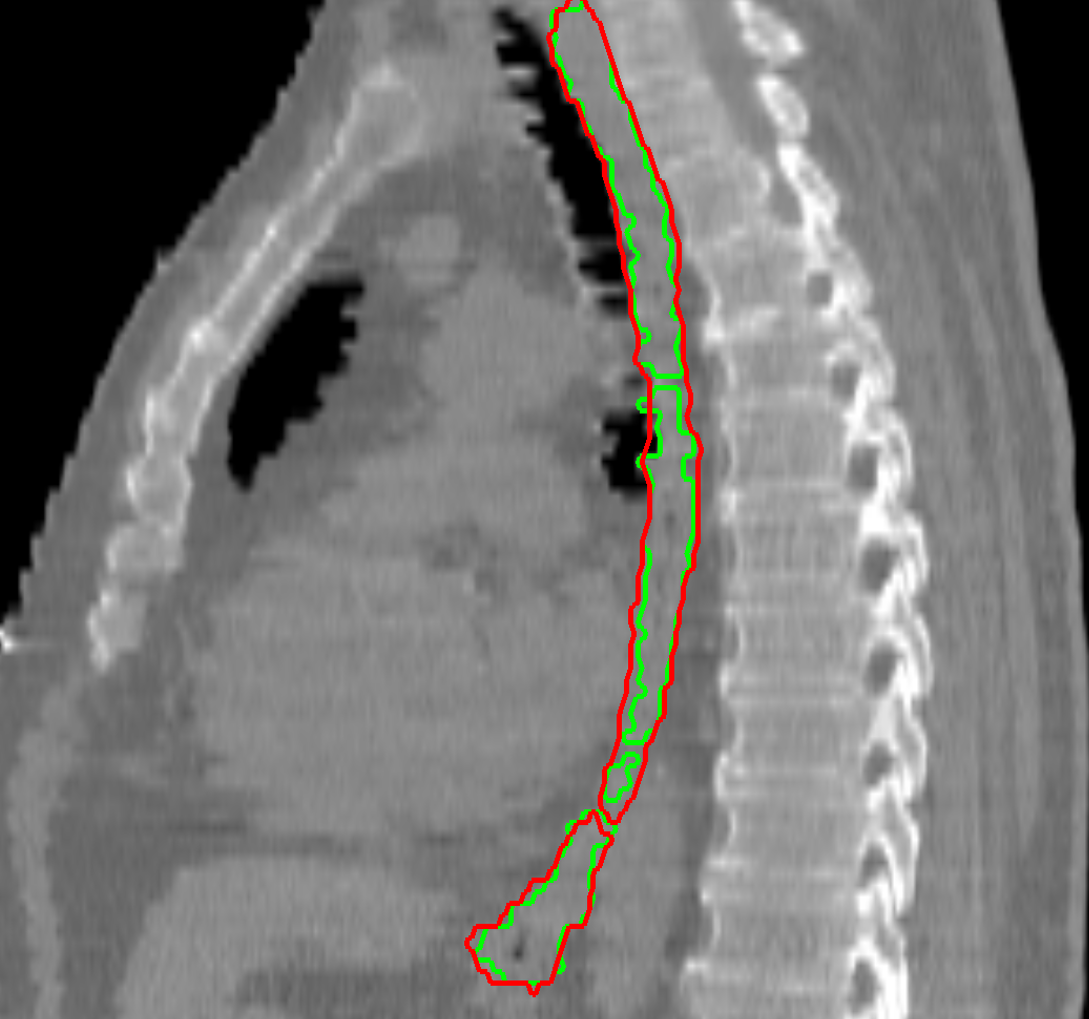} \\
        c) clinical case 20 \\ sagittal view}
               
        }
        
       \mbox{
      \shortstack{
        \includegraphics[width=0.3\linewidth]{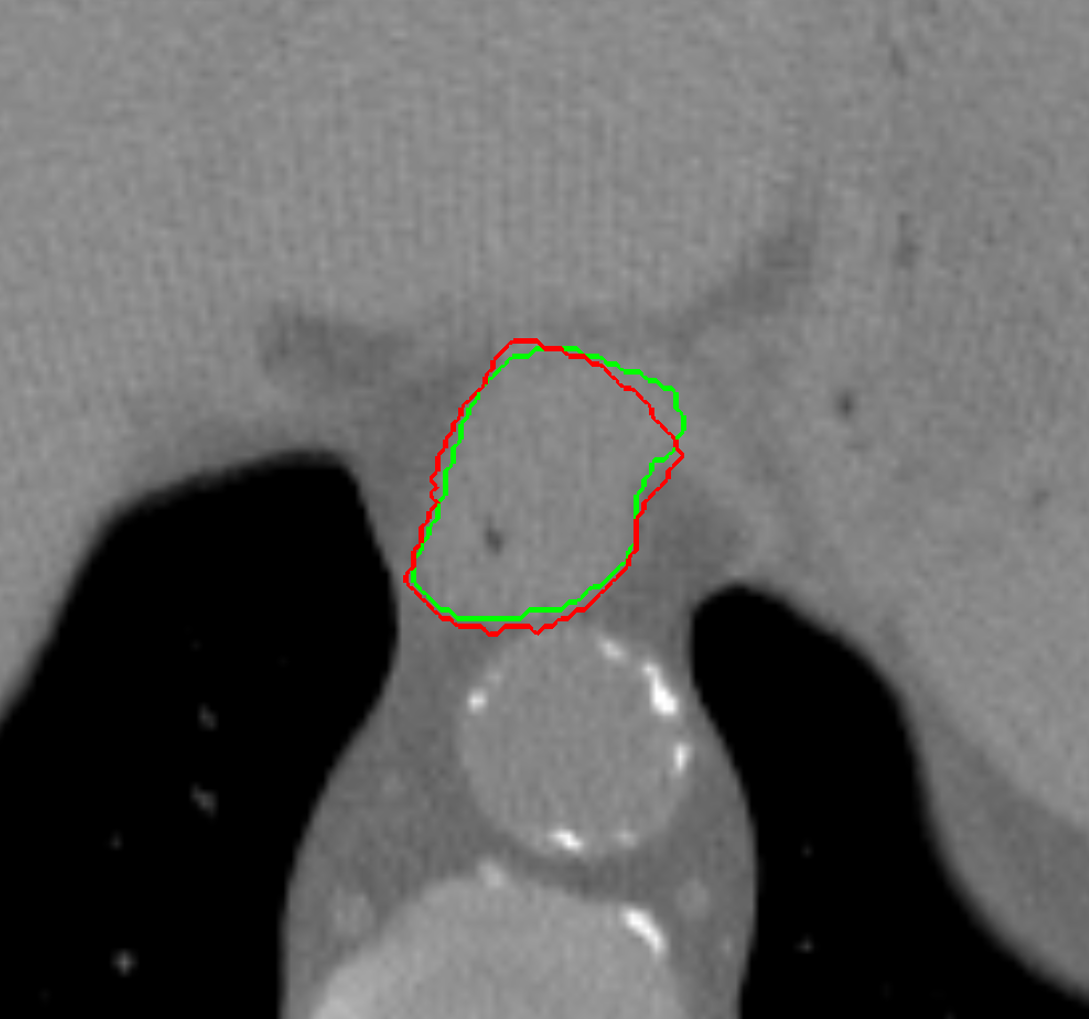} \\
        d) clinical case 09 \\ axial view
        }
        \hspace{-1.5 mm}
         
      \shortstack{     
        \includegraphics[width=0.3\linewidth]{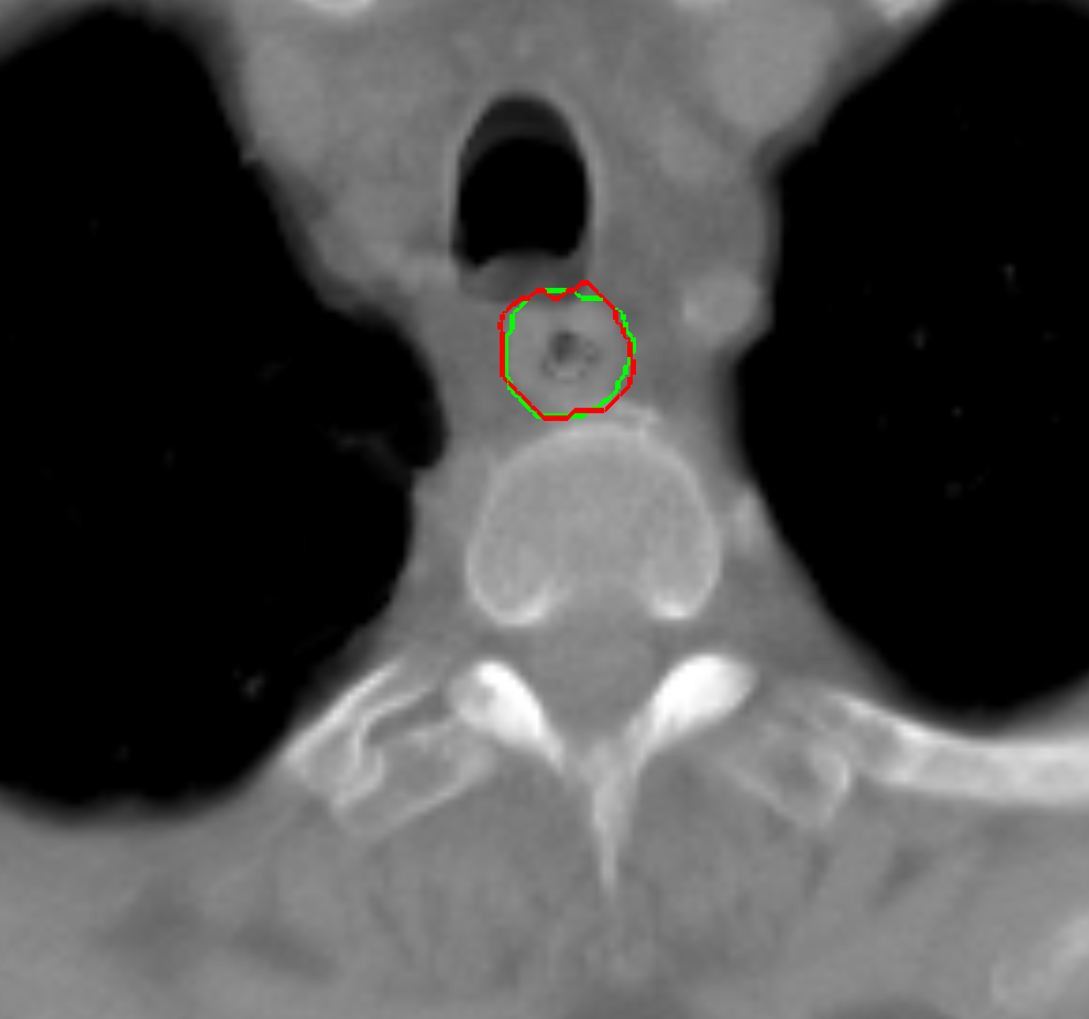} \\
        e) clinical case 13 \\ axial view}
        \hspace{-1.5 mm}

        \shortstack{
        \includegraphics[width=0.3\linewidth]{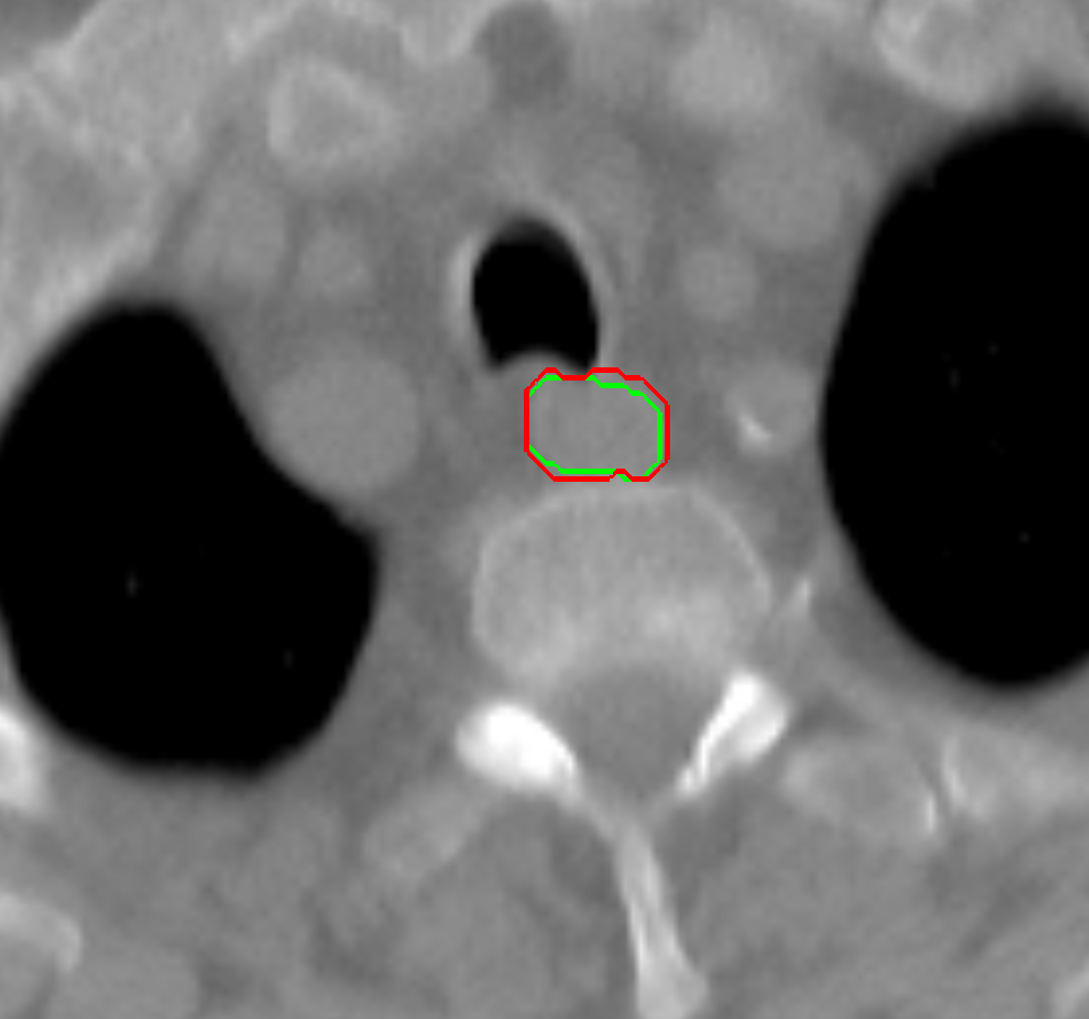} \\
        f) clinical case 20 \\ axial view}
               
        }
        
        \mbox{
      \shortstack{
        \includegraphics[width=0.3\linewidth]{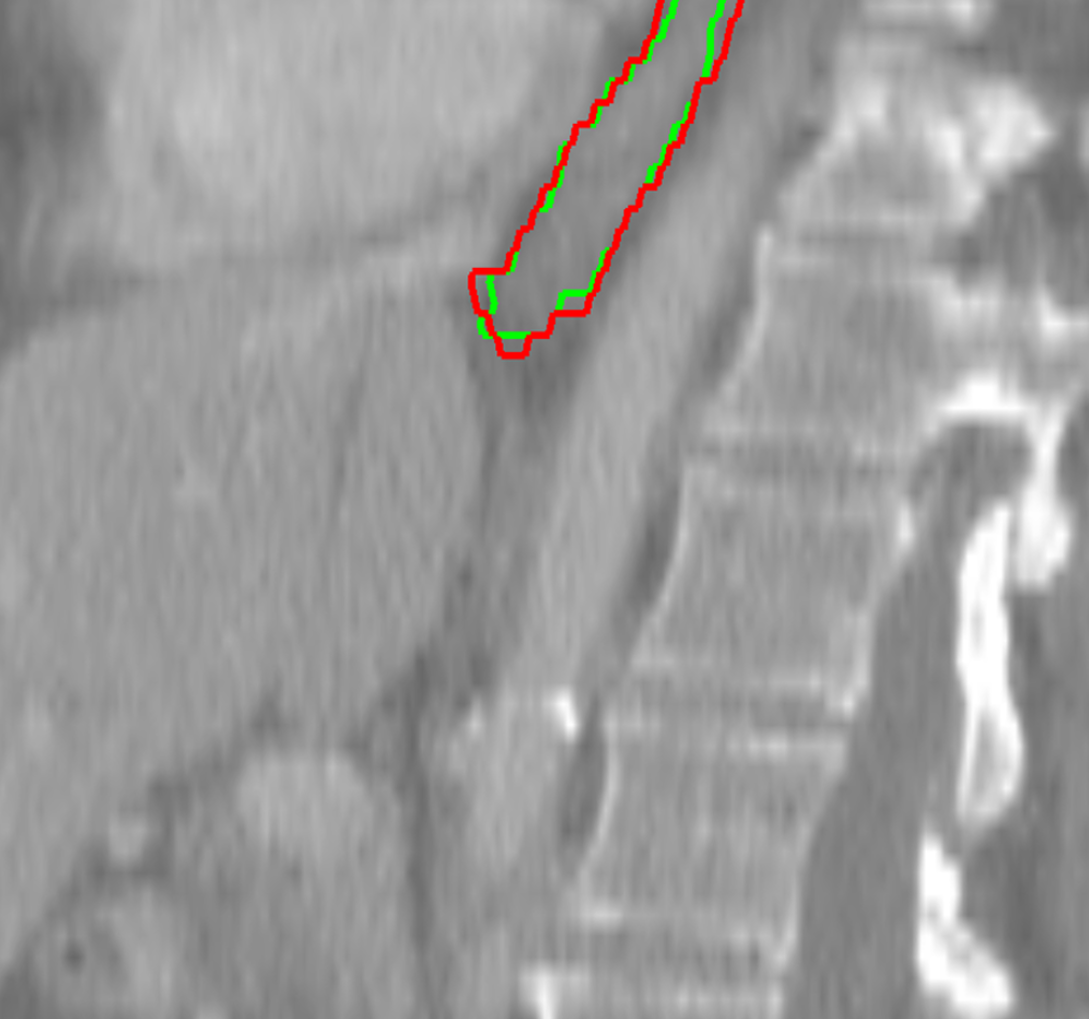} \\
        g) public case 01 \\ sagittal view
        }
        \hspace{-1.5 mm}
         
      \shortstack{     
        \includegraphics[width=0.3\linewidth]{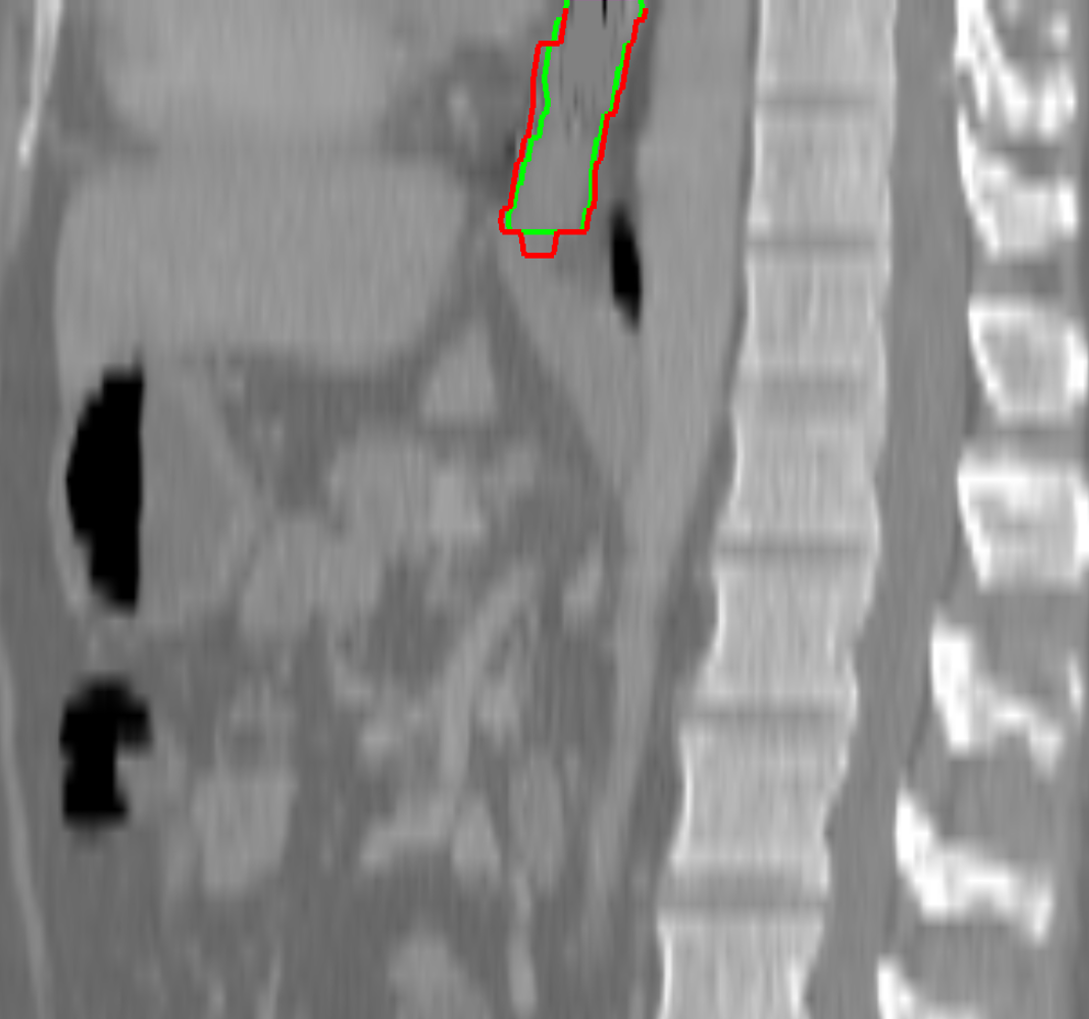} \\
        h) public case 12 \\ sagittal view}
        \hspace{-1.5 mm}

        \shortstack{
        \includegraphics[width=0.3\linewidth]{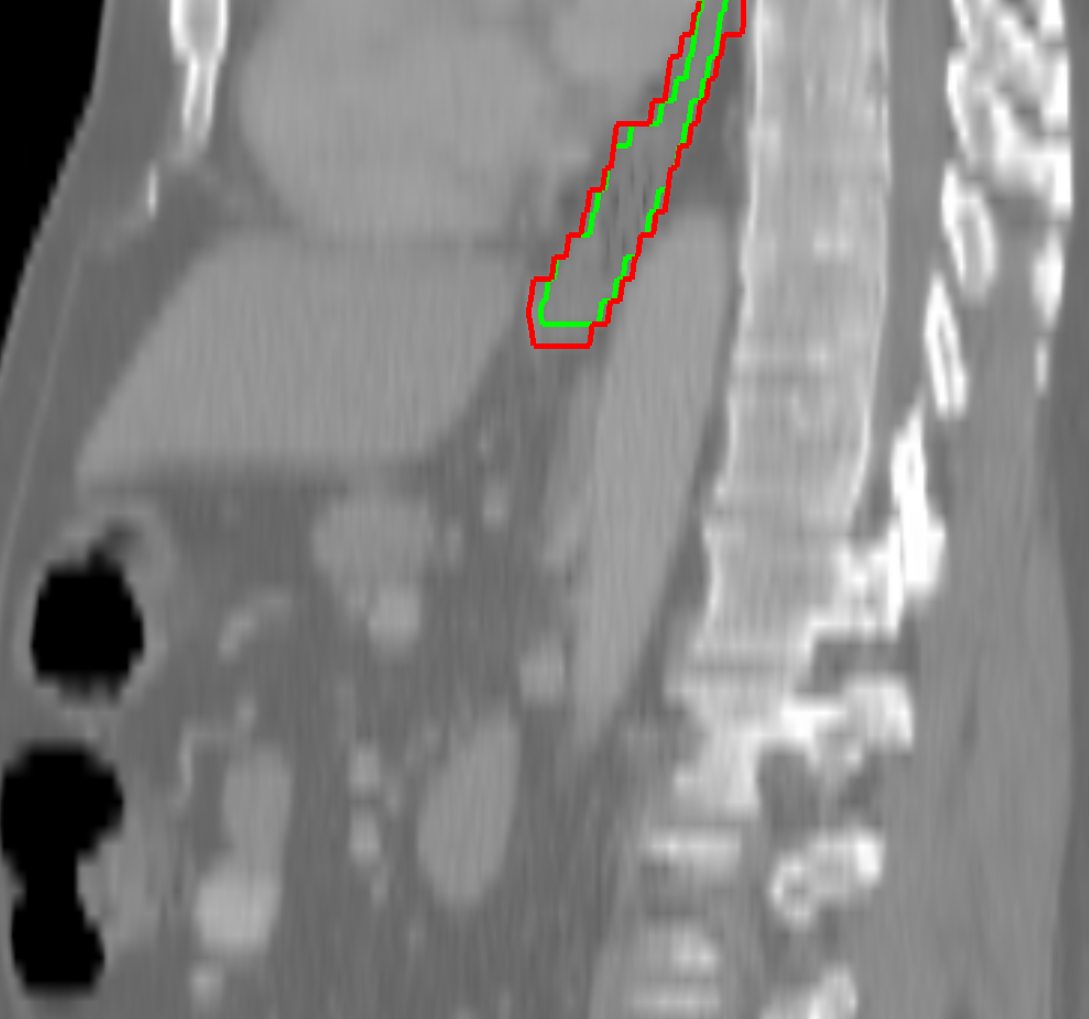} \\
        i) public case 24 \\ sagittal view}
               
        }
        
        \mbox{
      \shortstack{
        \includegraphics[width=0.3\linewidth]{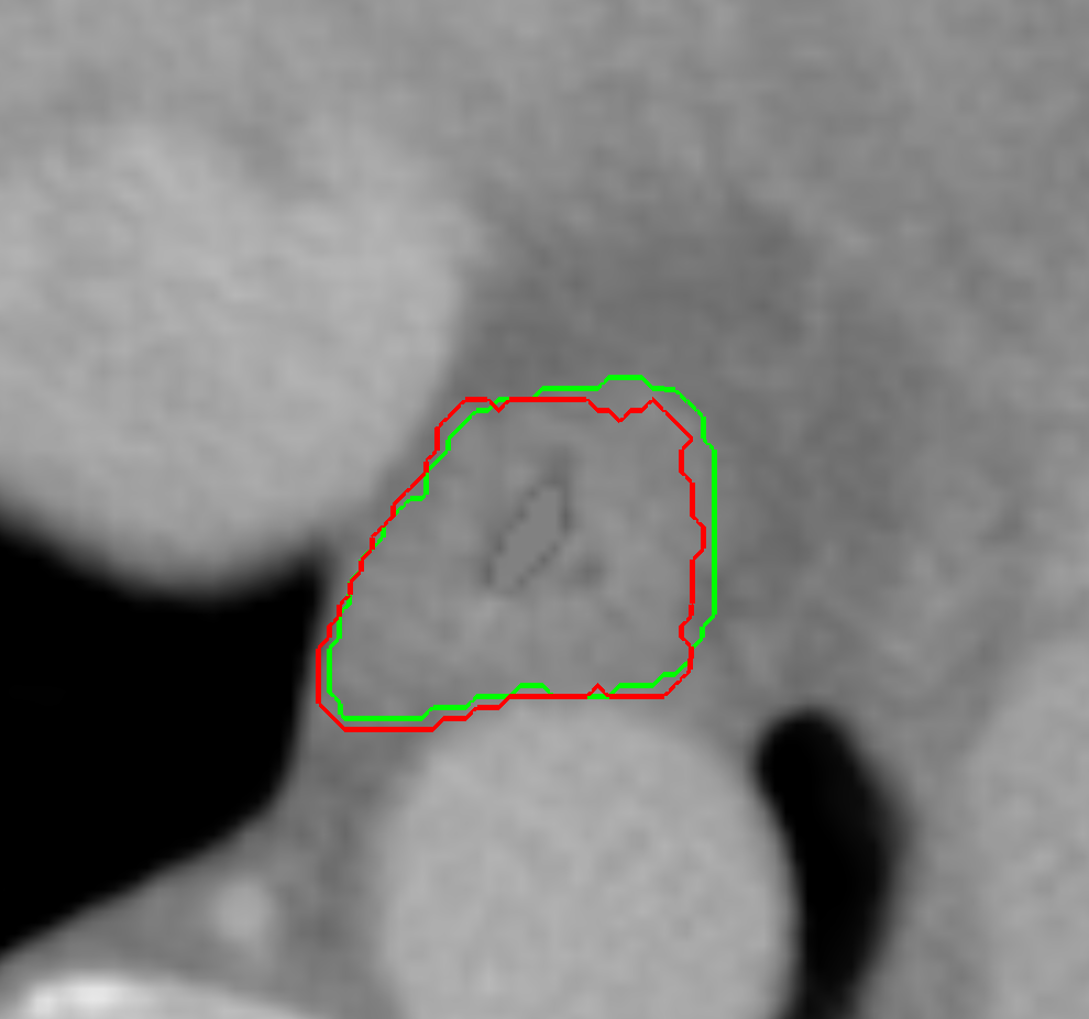} \\
        j) public case 01 \\ axial view
        }
        \hspace{-1.5 mm}
         
      \shortstack{     
        \includegraphics[width=0.3\linewidth]{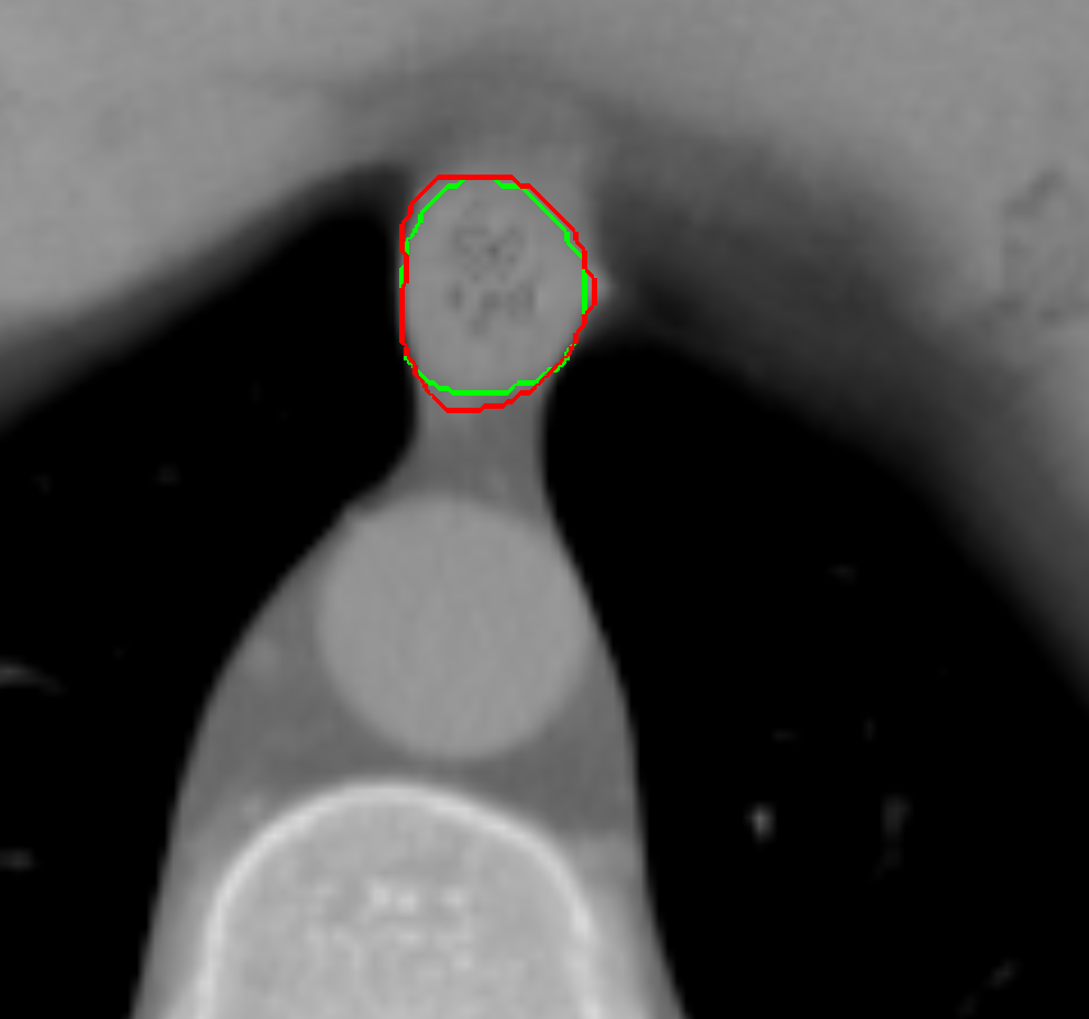} \\
        k) public case 12 \\ axial view}
        \hspace{-1.5 mm}

        \shortstack{
        \includegraphics[width=0.3\linewidth]{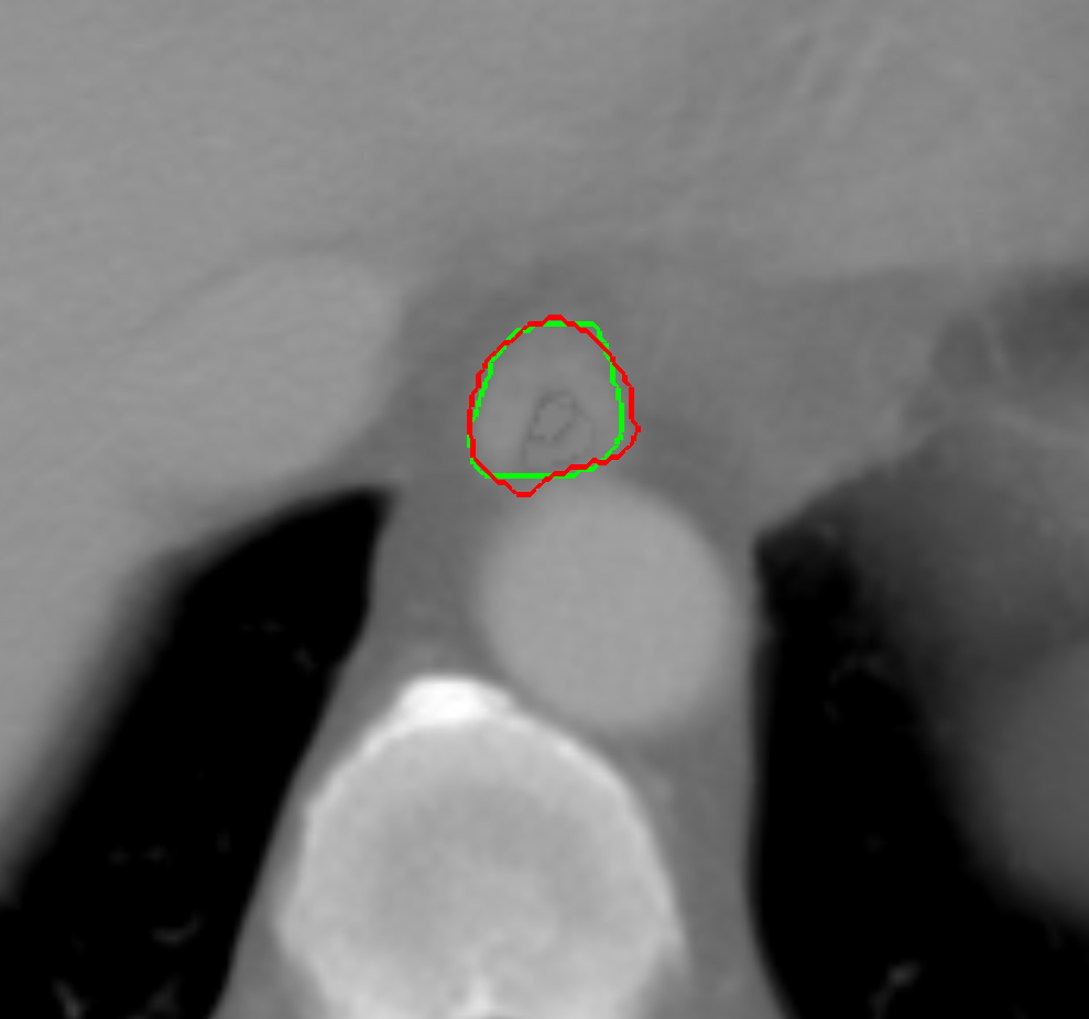} \\
        l) public case 24 \\ axial view}
               
        }
        \caption{Contouring results for 6 cases in different views. The green line indicates the reference standard, the red line depicts the contour generated by the proposed algorithm.}
\label{fig:results}
\end{center}        
\end{figure}

Visual results of our algorithm compared to the reference contour for 6 cases in axial and sagittal view are depicted in Fig. \ref{fig:results}. From the sagittal views it can be observed that manual contours are not as smooth as the automatic contours, which could be due to the breathing artifacts. The use of convolutions and post-processing in 3D aids at processing information of the 3D context through several slices, which is reflected in the smoother automatic contours shown in these images. On the other hand, from the axial views we can realize that, despite of the similarity with respect to neighboring tissues, and heterogeneity of the inner region of the esophagus, our automatic system provides contours that are comparable with the reference standard.



Segmentation was achieved in 90-100 seconds, on average, in a whole CT scan. In particular, CNN probability map calculation took between 40-60 seconds, ACM probability map calculation 30 seconds, CT probability values 7 seconds and generating the final contour by processing all probability maps with the RW 10 seconds on average.

\section{Discussion}

\begin{figure}[t!]
     \begin{center}
     \mbox{
      \shortstack{
        \includegraphics[width=0.48\linewidth]{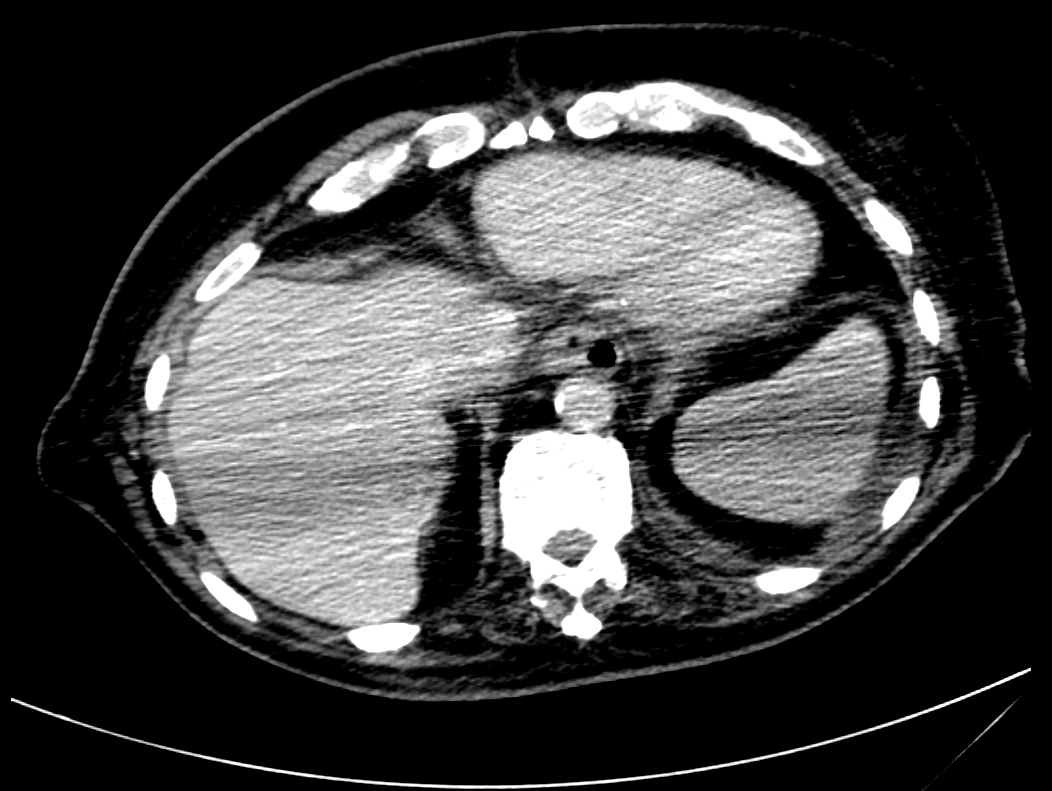} \\
        a) CT slice with artifacts of public case 7
        }
        \hspace{-1.5 mm}
         
      \shortstack{     
        \includegraphics[width=0.48\linewidth]{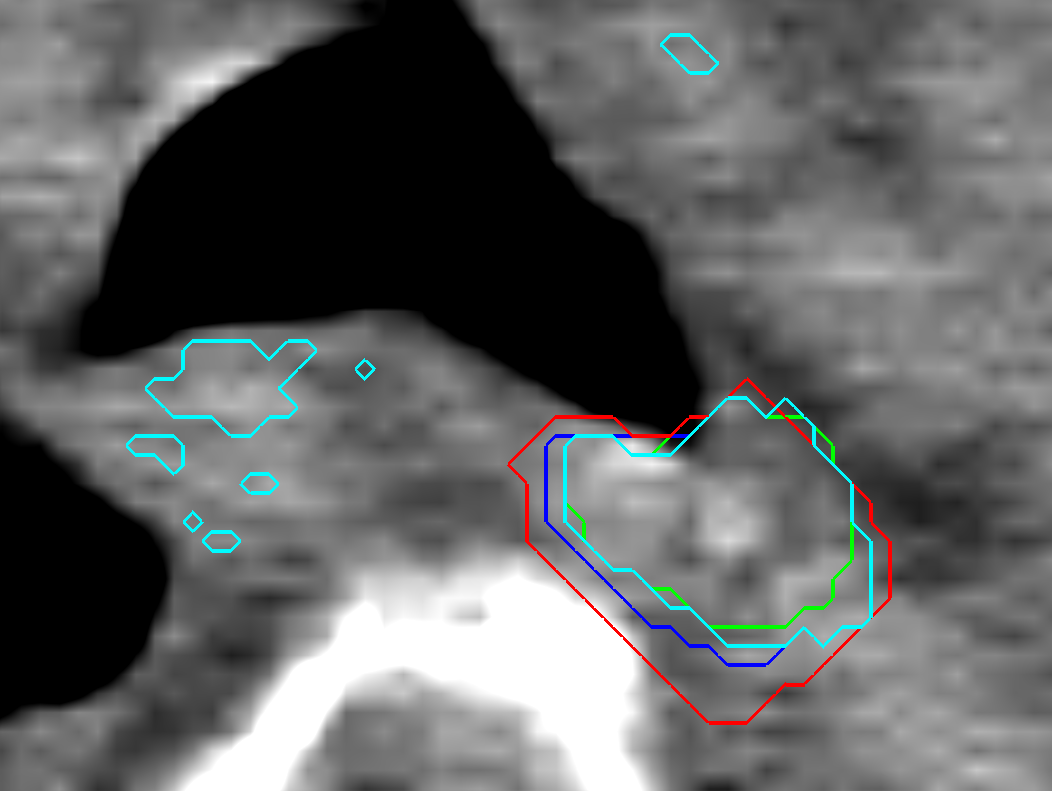} \\
        b) Segmentation results for clinical case 10}      
        }
\caption{In a) a CT slice of case 7 with intense artifacts caused by a placing of the arms next to thorax and abdomen is shown. Fig. b) shows one slice of clinical case 10 for which the RW processing decreased the segmentation accuracy (reference contour: green, RW contour: red, CNN 50 \% threshold: cyan, CNN 50 \% threshold and morphological closing: blue).}
\label{fig:discussion}
\end{center}        
\end{figure}

\begin{figure}[t!]
\begin{center}
     \mbox{

        \includegraphics[width=0.3\linewidth]{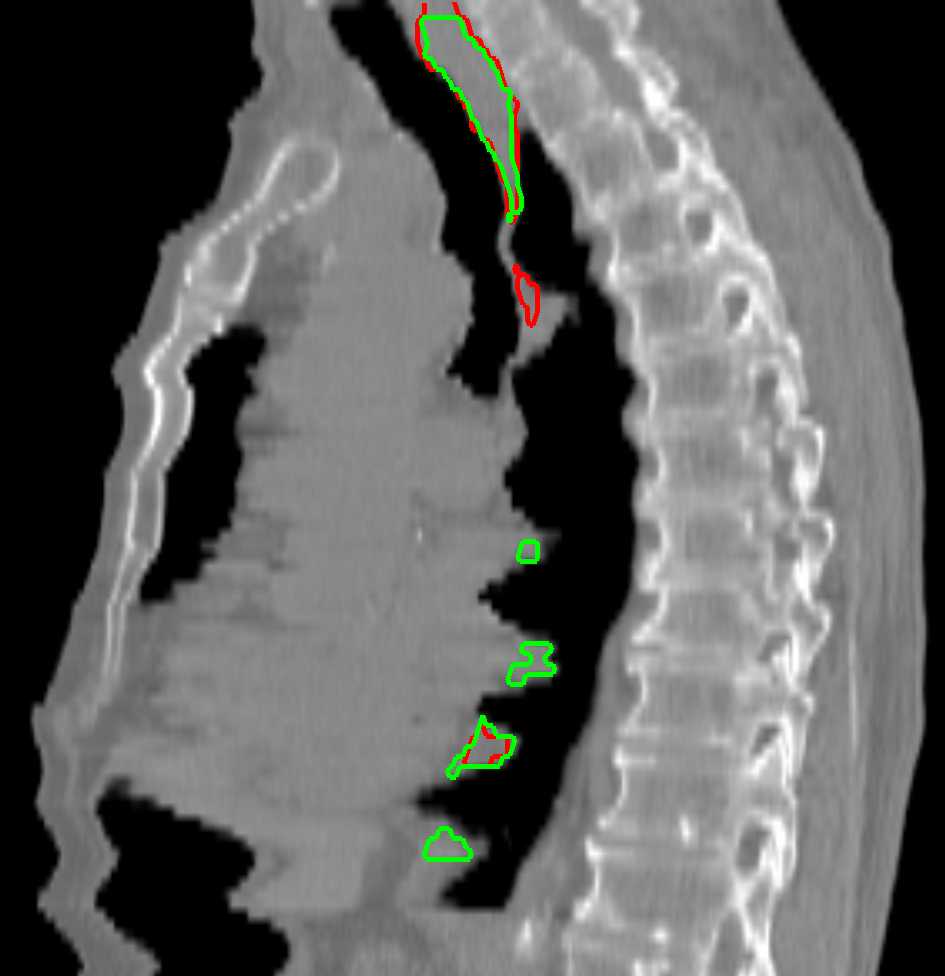}  
        \hspace{2 mm}

        \includegraphics[width=0.3\linewidth]{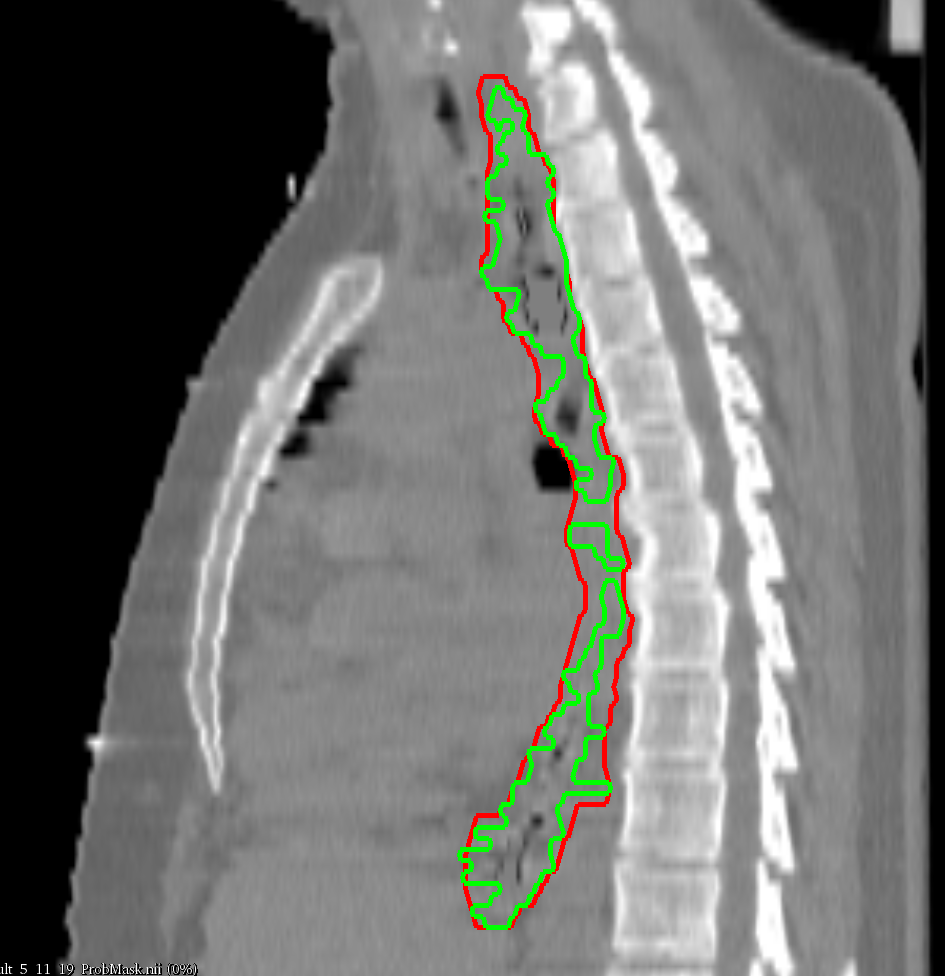}
  
        \hspace{2 mm}

        \includegraphics[width=0.3\linewidth]{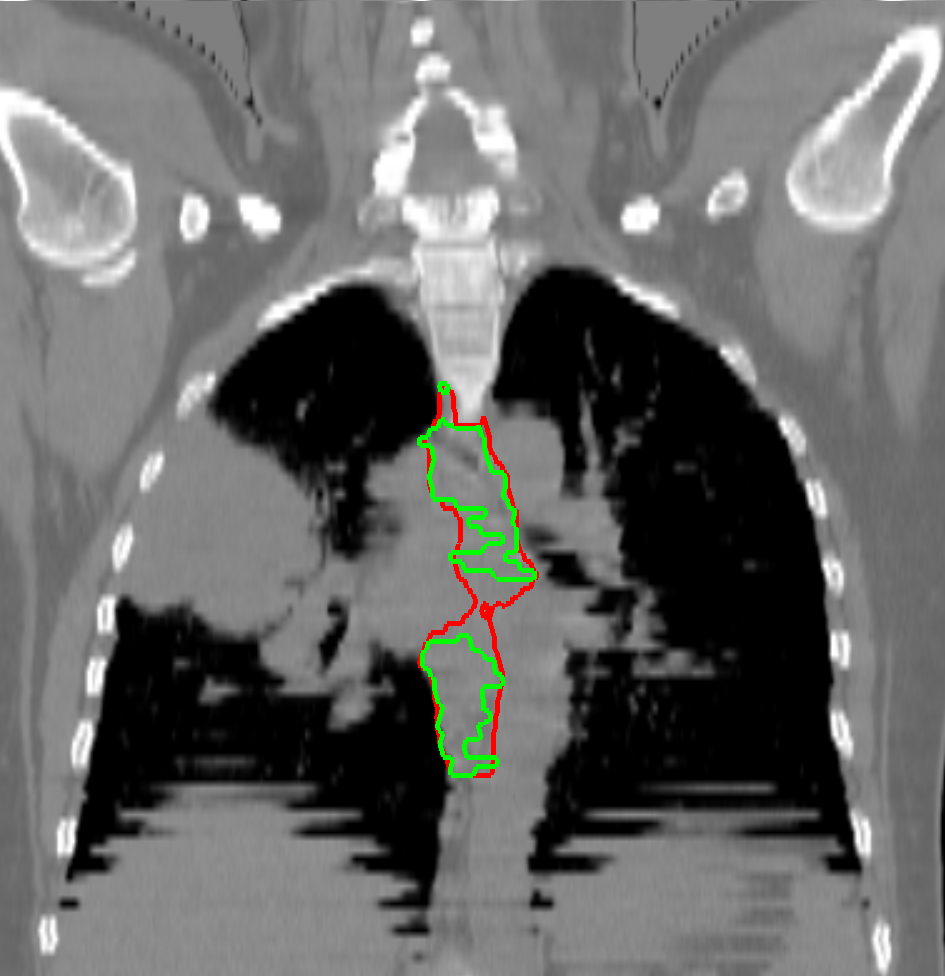}
             
        }
        \caption{Respiratory motion artifacts with reference contour (green) and generated contour (red).}
\label{fig:motionARtefacts}
\end{center}        
\end{figure}

We explored the feasibility of driving a random walk (RW) algorithm \cite{gradyML} with a CNN and its application to CT esophagus segmentation. Our results prove that a CNN is able to contour the esophagus almost to a perfect extent in some cases (DSC $>$ 0.9 for clinical cases 09 and 10). However, one can also note that the DSC values for the CNN strongly vary over all cases. Adding a RW for post processing made the performance more homogeneous and, as a whole, increased all figures of merit. The reason why CNN + RW worsened results for some cases is that for those cases the CNN 50 \% outline fitted the reference contour already very well and showed very few false positive spots. The RW post processing then enlarged the contour, which resulted in worse figures of merit (an example can be seen in Fig. \ref{fig:discussion}b). The enlargement was due to ''jumps'' of the depicted esophagus from one slice to the consecutive ones. As the RW algorithm considers the neighborhood of voxels in all directions it enclosed non-esophageal tissue in those ''jump'' areas. Examples can be seen in Fig. \ref{fig:motionARtefacts} and Fig. \ref{fig:results} a--c.

The ''jumps'' in the clinical cases arose because patients were freely breathing during CT acquisition. The respiratory motion of the thorax during acquisition caused numerous image artifacts. Some examples are depicted in Fig. \ref{fig:motionARtefacts}. Considering the bad image quality due to respiration, emphasizes the good performance of the proposed algorithm even more. Also the big variations in the CNN results without RW post processing may be due to the rather small training datasets and/or to the respiratory motion artifacts. Training and testing the algorithm again on gated or average CTs could probably increase its accuracy.

For public dataset 7 the CNN resulted in a probability map with almost all values close to zero. We suggest that this arises from the strong artifacts (see Fig. \ref{fig:discussion}a) only visible in this case. The stripe shaped artifacts change the texture of all visible organs and hence the results of the convolution filter. As this case is the only one in the dataset with such artifacts the CNN was not able to learn how to classify the differently looking structures correctly. The origin of the artifact seems to be the arms of the patients lying parallel to the thorax and abdomen in the reconstructed field of view (FOV). As far as it can be assumed taking into account the often very limited FOV of the public datasets, all other patients placed their arms in a different position.


Previous methods, which are presented in table \ref{table:Results}, mostly require diverse levels of user interaction. Although the proposed method does not outperform the best semi-automatic approach \cite{rousson2006probabilistic}, it has the advantage of being fully automatic. This feature is particularly important if segmentation of massive data is required. Regarding the comparison with the only automatic method published up to date to segment the esophagus, our approach obtains slightly better volume overlapping values while reduces both mean and Hausdorff distances between automatic and reference contours. Taking into account these results we can thereby claim that up to date the proposed approach represents the state-of-the-art for the problem of automatic esophagus segmentation. 

Our promising results open the perspective to perform further evaluations to investigate the impact of our automatic contours on radiotherapy planning. Published studies showed that esophageal motion during one treatment fraction and between two or more fractions is a non negligibly issue. Due to respiration and cardiac motion intra fractional movement of the esophagus between 5 and 10 mm was reported\cite{cohenEsophagusMotion, palmerEsophagusMotionCardiac}, even though it can be up to 15 mm\cite{palmerEsophagusMotionCardiac,yamashitaEsophagusMotionCBCT}. Additionally, it was shown that the motion is highly patient and location dependent\cite{palmerEsophagusMotionCardiac,yamashitaEsophagusMotionCBCT}. Therefore, an expansion and evaluation of our algorithm to 4D time gated CT and cone beam CT is intended. On the one hand to provide an automatic tool to measure patient specific esophageal motion for treatment planning and on the other hand to offer a fast and reliable method to account for inter fractional motion and setup errors to enable adaptive planning. To do that, one option might be to generate segmentations in 3D volumes across different time frames, and apply a conditional random field as a post-processing step to impose consistency in time.

\section{Conclusion}
In this work we present a fully automatic algorithm for contouring the esophagus on CT. Our algorithm first generates an accurate estimation of the esophagus by employing a CNN. In a second step the estimation is refined to the final contour by using an active contour model and a random walker. One of the main advantages of our CNN are the 3D convolution kernels, which allow a fully exploitation of the 3D spatial CT information. Compared to other automatic algorithms our method shows superior spatial overlap and similarity in shape with the reference contours. The very promising results (DSC: 0.76, ASSD: 1.36, HD: 11.68) suggest further evaluations and to investigate a possible utilization in the context of radiotherapy planning.

\textbf{Acknowledgments:} This work is supported by the National Science and Engineering Research Council of Canada (NSERC), discovery grant program, and by the ETS Research Chair on Artificial Intelligence in Medical Imaging.

\textbf{Disclosure of Conflicts of Interest:} The authors have no relevant conflicts of interest to disclose.


\begin{thebibliography}{32}%
\makeatletter
\providecommand \@ifxundefined [1]{%
 \@ifx{#1\undefined}
}%
\providecommand \@ifnum [1]{%
 \ifnum #1\expandafter \@firstoftwo
 \else \expandafter \@secondoftwo
 \fi
}%
\providecommand \@ifx [1]{%
 \ifx #1\expandafter \@firstoftwo
 \else \expandafter \@secondoftwo
 \fi
}%
\providecommand \natexlab [1]{#1}%
\providecommand \enquote  [1]{``#1''}%
\providecommand \bibnamefont  [1]{#1}%
\providecommand \bibfnamefont [1]{#1}%
\providecommand \citenamefont [1]{#1}%
\providecommand \href@noop [0]{\@secondoftwo}%
\providecommand \href [0]{\begingroup \@sanitize@url \@href}%
\providecommand \@href[1]{\@@startlink{#1}\@@href}%
\providecommand \@@href[1]{\endgroup#1\@@endlink}%
\providecommand \@sanitize@url [0]{\catcode `\\12\catcode `\$12\catcode
  `\&12\catcode `\#12\catcode `\^12\catcode `\_12\catcode `\%12\relax}%
\providecommand \@@startlink[1]{}%
\providecommand \@@endlink[0]{}%
\providecommand \url  [0]{\begingroup\@sanitize@url \@url }%
\providecommand \@url [1]{\endgroup\@href {#1}{\urlprefix }}%
\providecommand \urlprefix  [0]{URL }%
\providecommand \Eprint [0]{\href }%
\providecommand \doibase [0]{http://dx.doi.org/}%
\providecommand \selectlanguage [0]{\@gobble}%
\providecommand \bibinfo  [0]{\@secondoftwo}%
\providecommand \bibfield  [0]{\@secondoftwo}%
\providecommand \translation [1]{[#1]}%
\providecommand \BibitemOpen [0]{}%
\providecommand \bibitemStop [0]{}%
\providecommand \bibitemNoStop [0]{.\EOS\space}%
\providecommand \EOS [0]{\spacefactor3000\relax}%
\providecommand \BibitemShut  [1]{\csname bibitem#1\endcsname}%
\let\auto@bib@innerbib\@empty
\bibitem [{\citenamefont {Collier}\ \emph {et~al.}(2003)\citenamefont
  {Collier}, \citenamefont {Burnett}, \citenamefont {Amin}, \citenamefont
  {Bilton}, \citenamefont {Brooks}, \citenamefont {Ryan}, \citenamefont
  {Roniger}, \citenamefont {Tran},\ and\ \citenamefont
  {Starkschall}}]{collier2003}%
  \BibitemOpen
  \bibfield  {author} {\bibinfo {author} {\bibfnamefont {D.~C.}\ \bibnamefont
  {Collier}}, \bibinfo {author} {\bibfnamefont {S.~S.}\ \bibnamefont
  {Burnett}}, \bibinfo {author} {\bibfnamefont {M.}~\bibnamefont {Amin}},
  \bibinfo {author} {\bibfnamefont {S.}~\bibnamefont {Bilton}}, \bibinfo
  {author} {\bibfnamefont {C.}~\bibnamefont {Brooks}}, \bibinfo {author}
  {\bibfnamefont {A.}~\bibnamefont {Ryan}}, \bibinfo {author} {\bibfnamefont
  {D.}~\bibnamefont {Roniger}}, \bibinfo {author} {\bibfnamefont
  {D.}~\bibnamefont {Tran}}, \ and\ \bibinfo {author} {\bibfnamefont
  {G.}~\bibnamefont {Starkschall}},\ }\bibfield  {title} {\enquote {\bibinfo
  {title} {Assessment of consistency in contouring of normal-tissue anatomic
  structures},}\ }\href@noop {} {\bibfield  {journal} {\bibinfo  {journal}
  {Journal of Applied Clinical Medical Physics}\ }\textbf {\bibinfo {volume}
  {4}},\ \bibinfo {pages} {17--24} (\bibinfo {year} {2003})}\BibitemShut
  {NoStop}%
\bibitem [{\citenamefont {Heimann}\ and\ \citenamefont
  {Meinzer}(2009)}]{heimann2009statistical}%
  \BibitemOpen
  \bibfield  {author} {\bibinfo {author} {\bibfnamefont {T.}~\bibnamefont
  {Heimann}}\ and\ \bibinfo {author} {\bibfnamefont {H.-P.}\ \bibnamefont
  {Meinzer}},\ }\bibfield  {title} {\enquote {\bibinfo {title} {Statistical
  shape models for 3d medical image segmentation: a review},}\ }\href@noop {}
  {\bibfield  {journal} {\bibinfo  {journal} {Medical image analysis}\ }\textbf
  {\bibinfo {volume} {13}},\ \bibinfo {pages} {543--563} (\bibinfo {year}
  {2009})}\BibitemShut {NoStop}%
\bibitem [{\citenamefont {Ragan}\ \emph {et~al.}(2005)\citenamefont {Ragan},
  \citenamefont {Starkschall}, \citenamefont {McNutt}, \citenamefont {Kaus},
  \citenamefont {Guerrero},\ and\ \citenamefont
  {Stevens}}]{ragan2005semiautomated}%
  \BibitemOpen
  \bibfield  {author} {\bibinfo {author} {\bibfnamefont {D.}~\bibnamefont
  {Ragan}}, \bibinfo {author} {\bibfnamefont {G.}~\bibnamefont {Starkschall}},
  \bibinfo {author} {\bibfnamefont {T.}~\bibnamefont {McNutt}}, \bibinfo
  {author} {\bibfnamefont {M.}~\bibnamefont {Kaus}}, \bibinfo {author}
  {\bibfnamefont {T.}~\bibnamefont {Guerrero}}, \ and\ \bibinfo {author}
  {\bibfnamefont {C.~W.}\ \bibnamefont {Stevens}},\ }\bibfield  {title}
  {\enquote {\bibinfo {title} {Semiautomated four-dimensional computed
  tomography segmentation using deformable models},}\ }\href@noop {} {\bibfield
   {journal} {\bibinfo  {journal} {Medical physics}\ }\textbf {\bibinfo
  {volume} {32}},\ \bibinfo {pages} {2254--2261} (\bibinfo {year}
  {2005})}\BibitemShut {NoStop}%
\bibitem [{\citenamefont {Dolz}\ \emph {et~al.}(2016)\citenamefont {Dolz},
  \citenamefont {Kiri{\c{s}}li}, \citenamefont {Fechter}, \citenamefont
  {Karnitzki}, \citenamefont {Oehlke}, \citenamefont {Nestle}, \citenamefont
  {Vermandel},\ and\ \citenamefont {Massoptier}}]{dolz2016interactive}%
  \BibitemOpen
  \bibfield  {author} {\bibinfo {author} {\bibfnamefont {J.}~\bibnamefont
  {Dolz}}, \bibinfo {author} {\bibfnamefont {H.}~\bibnamefont {Kiri{\c{s}}li}},
  \bibinfo {author} {\bibfnamefont {T.}~\bibnamefont {Fechter}}, \bibinfo
  {author} {\bibfnamefont {S.}~\bibnamefont {Karnitzki}}, \bibinfo {author}
  {\bibfnamefont {O.}~\bibnamefont {Oehlke}}, \bibinfo {author} {\bibfnamefont
  {U.}~\bibnamefont {Nestle}}, \bibinfo {author} {\bibfnamefont
  {M.}~\bibnamefont {Vermandel}}, \ and\ \bibinfo {author} {\bibfnamefont
  {L.}~\bibnamefont {Massoptier}},\ }\bibfield  {title} {\enquote {\bibinfo
  {title} {Interactive contour delineation of organs at risk in radiotherapy:
  Clinical evaluation on nsclc patients.}}\ }\href@noop {} {\bibfield
  {journal} {\bibinfo  {journal} {Medical physics}\ }\textbf {\bibinfo {volume}
  {43}},\ \bibinfo {pages} {2569--2569} (\bibinfo {year} {2016})}\BibitemShut
  {NoStop}%
\bibitem [{\citenamefont {Rousson}\ \emph {et~al.}(2006)\citenamefont
  {Rousson}, \citenamefont {Bai}, \citenamefont {Xu},\ and\ \citenamefont
  {Sauer}}]{rousson2006probabilistic}%
  \BibitemOpen
  \bibfield  {author} {\bibinfo {author} {\bibfnamefont {M.}~\bibnamefont
  {Rousson}}, \bibinfo {author} {\bibfnamefont {Y.}~\bibnamefont {Bai}},
  \bibinfo {author} {\bibfnamefont {C.}~\bibnamefont {Xu}}, \ and\ \bibinfo
  {author} {\bibfnamefont {F.}~\bibnamefont {Sauer}},\ }\bibfield  {title}
  {\enquote {\bibinfo {title} {Probabilistic minimal path for automated
  esophagus segmentation},}\ }in\ \href@noop {} {\emph {\bibinfo {booktitle}
  {Medical Imaging}}}\ (\bibinfo {organization} {International Society for
  Optics and Photonics},\ \bibinfo {year} {2006})\ pp.\ \bibinfo {pages}
  {614449--614449}\BibitemShut {NoStop}%
\bibitem [{\citenamefont {Grosgeorge}\ \emph {et~al.}(2013)\citenamefont
  {Grosgeorge}, \citenamefont {Petitjean}, \citenamefont {Dubray},\ and\
  \citenamefont {Ruan}}]{grosgeorge2013esophagus}%
  \BibitemOpen
  \bibfield  {author} {\bibinfo {author} {\bibfnamefont {D.}~\bibnamefont
  {Grosgeorge}}, \bibinfo {author} {\bibfnamefont {C.}~\bibnamefont
  {Petitjean}}, \bibinfo {author} {\bibfnamefont {B.}~\bibnamefont {Dubray}}, \
  and\ \bibinfo {author} {\bibfnamefont {S.}~\bibnamefont {Ruan}},\ }\bibfield
  {title} {\enquote {\bibinfo {title} {Esophagus segmentation from 3d ct data
  using skeleton prior-based graph cut},}\ }\href@noop {} {\bibfield  {journal}
  {\bibinfo  {journal} {Computational and mathematical methods in medicine}\
  }\textbf {\bibinfo {volume} {2013}} (\bibinfo {year} {2013})}\BibitemShut
  {NoStop}%
\bibitem [{\citenamefont {Feulner}\ \emph {et~al.}(2009)\citenamefont
  {Feulner}, \citenamefont {Zhou}, \citenamefont {Cavallaro}, \citenamefont
  {Seifert}, \citenamefont {Hornegger},\ and\ \citenamefont
  {Comaniciu}}]{feulner2009fast}%
  \BibitemOpen
  \bibfield  {author} {\bibinfo {author} {\bibfnamefont {J.}~\bibnamefont
  {Feulner}}, \bibinfo {author} {\bibfnamefont {S.~K.}\ \bibnamefont {Zhou}},
  \bibinfo {author} {\bibfnamefont {A.}~\bibnamefont {Cavallaro}}, \bibinfo
  {author} {\bibfnamefont {S.}~\bibnamefont {Seifert}}, \bibinfo {author}
  {\bibfnamefont {J.}~\bibnamefont {Hornegger}}, \ and\ \bibinfo {author}
  {\bibfnamefont {D.}~\bibnamefont {Comaniciu}},\ }\bibfield  {title} {\enquote
  {\bibinfo {title} {Fast automatic segmentation of the esophagus from 3d ct
  data using a probabilistic model},}\ }in\ \href@noop {} {\emph {\bibinfo
  {booktitle} {International Conference on Medical Image Computing and
  Computer-Assisted Intervention}}}\ (\bibinfo {organization} {Springer},\
  \bibinfo {year} {2009})\ pp.\ \bibinfo {pages} {255--262}\BibitemShut
  {NoStop}%
\bibitem [{\citenamefont {Fieselmann}\ \emph {et~al.}(2008)\citenamefont
  {Fieselmann}, \citenamefont {Lautenschl{\"a}ger}, \citenamefont {Deinzer},
  \citenamefont {John},\ and\ \citenamefont {Poppe}}]{fieselmann2008esophagus}%
  \BibitemOpen
  \bibfield  {author} {\bibinfo {author} {\bibfnamefont {A.}~\bibnamefont
  {Fieselmann}}, \bibinfo {author} {\bibfnamefont {S.}~\bibnamefont
  {Lautenschl{\"a}ger}}, \bibinfo {author} {\bibfnamefont {F.}~\bibnamefont
  {Deinzer}}, \bibinfo {author} {\bibfnamefont {M.}~\bibnamefont {John}}, \
  and\ \bibinfo {author} {\bibfnamefont {B.}~\bibnamefont {Poppe}},\ }\bibfield
   {title} {\enquote {\bibinfo {title} {Esophagus segmentation by
  spatially-constrained shape interpolation},}\ }in\ \href@noop {} {\emph
  {\bibinfo {booktitle} {Bildverarbeitung f{\"u}r die Medizin 2008}}}\
  (\bibinfo  {publisher} {Springer},\ \bibinfo {year} {2008})\ pp.\ \bibinfo
  {pages} {247--251}\BibitemShut {NoStop}%
\bibitem [{\citenamefont {Kurugol}\ \emph {et~al.}(2011)\citenamefont
  {Kurugol}, \citenamefont {Bas}, \citenamefont {Erdogmus}, \citenamefont {Dy},
  \citenamefont {Sharp},\ and\ \citenamefont {Brooks}}]{kurugol2011centerline}%
  \BibitemOpen
  \bibfield  {author} {\bibinfo {author} {\bibfnamefont {S.}~\bibnamefont
  {Kurugol}}, \bibinfo {author} {\bibfnamefont {E.}~\bibnamefont {Bas}},
  \bibinfo {author} {\bibfnamefont {D.}~\bibnamefont {Erdogmus}}, \bibinfo
  {author} {\bibfnamefont {J.~G.}\ \bibnamefont {Dy}}, \bibinfo {author}
  {\bibfnamefont {G.~C.}\ \bibnamefont {Sharp}}, \ and\ \bibinfo {author}
  {\bibfnamefont {D.~H.}\ \bibnamefont {Brooks}},\ }\bibfield  {title}
  {\enquote {\bibinfo {title} {Centerline extraction with principal curve
  tracing to improve 3d level set esophagus segmentation in ct images},}\ }in\
  \href@noop {} {\emph {\bibinfo {booktitle} {2011 Annual International
  Conference of the IEEE Engineering in Medicine and Biology Society}}}\
  (\bibinfo {organization} {IEEE},\ \bibinfo {year} {2011})\ pp.\ \bibinfo
  {pages} {3403--3406}\BibitemShut {NoStop}%
\bibitem [{\citenamefont {LeCun}\ \emph {et~al.}(1998)\citenamefont {LeCun},
  \citenamefont {Bottou}, \citenamefont {Bengio},\ and\ \citenamefont
  {Haffner}}]{lecun1998gradient}%
  \BibitemOpen
  \bibfield  {author} {\bibinfo {author} {\bibfnamefont {Y.}~\bibnamefont
  {LeCun}}, \bibinfo {author} {\bibfnamefont {L.}~\bibnamefont {Bottou}},
  \bibinfo {author} {\bibfnamefont {Y.}~\bibnamefont {Bengio}}, \ and\ \bibinfo
  {author} {\bibfnamefont {P.}~\bibnamefont {Haffner}},\ }\bibfield  {title}
  {\enquote {\bibinfo {title} {Gradient-based learning applied to document
  recognition},}\ }\href@noop {} {\bibfield  {journal} {\bibinfo  {journal}
  {Proceedings of the IEEE}\ }\textbf {\bibinfo {volume} {86}},\ \bibinfo
  {pages} {2278--2324} (\bibinfo {year} {1998})}\BibitemShut {NoStop}%
\bibitem [{\citenamefont {Litjens}\ \emph {et~al.}(2017)\citenamefont
  {Litjens}, \citenamefont {Kooi}, \citenamefont {Bejnordi}, \citenamefont
  {Setio}, \citenamefont {Ciompi}, \citenamefont {Ghafoorian}, \citenamefont
  {van~der Laak}, \citenamefont {van Ginneken},\ and\ \citenamefont
  {S{\'a}nchez}}]{litjens2017survey}%
  \BibitemOpen
  \bibfield  {author} {\bibinfo {author} {\bibfnamefont {G.}~\bibnamefont
  {Litjens}}, \bibinfo {author} {\bibfnamefont {T.}~\bibnamefont {Kooi}},
  \bibinfo {author} {\bibfnamefont {B.~E.}\ \bibnamefont {Bejnordi}}, \bibinfo
  {author} {\bibfnamefont {A.~A.~A.}\ \bibnamefont {Setio}}, \bibinfo {author}
  {\bibfnamefont {F.}~\bibnamefont {Ciompi}}, \bibinfo {author} {\bibfnamefont
  {M.}~\bibnamefont {Ghafoorian}}, \bibinfo {author} {\bibfnamefont {J.~A.}\
  \bibnamefont {van~der Laak}}, \bibinfo {author} {\bibfnamefont
  {B.}~\bibnamefont {van Ginneken}}, \ and\ \bibinfo {author} {\bibfnamefont
  {C.~I.}\ \bibnamefont {S{\'a}nchez}},\ }\bibfield  {title} {\enquote
  {\bibinfo {title} {A survey on deep learning in medical image analysis},}\
  }\href@noop {} {\bibfield  {journal} {\bibinfo  {journal} {arXiv preprint
  arXiv:1702.05747}\ } (\bibinfo {year} {2017})}\BibitemShut {NoStop}%
\bibitem [{\citenamefont {Long}, \citenamefont {Shelhamer},\ and\ \citenamefont
  {Darrell}(2015)}]{long2015fully}%
  \BibitemOpen
  \bibfield  {author} {\bibinfo {author} {\bibfnamefont {J.}~\bibnamefont
  {Long}}, \bibinfo {author} {\bibfnamefont {E.}~\bibnamefont {Shelhamer}}, \
  and\ \bibinfo {author} {\bibfnamefont {T.}~\bibnamefont {Darrell}},\
  }\bibfield  {title} {\enquote {\bibinfo {title} {Fully convolutional networks
  for semantic segmentation},}\ }in\ \href@noop {} {\emph {\bibinfo {booktitle}
  {Proceedings of the IEEE Conference on Computer Vision and Pattern
  Recognition}}}\ (\bibinfo {year} {2015})\ pp.\ \bibinfo {pages}
  {3431--3440}\BibitemShut {NoStop}%
\bibitem [{\citenamefont {Szegedy}\ \emph {et~al.}(2015)\citenamefont
  {Szegedy}, \citenamefont {Liu}, \citenamefont {Jia}, \citenamefont
  {Sermanet}, \citenamefont {Reed}, \citenamefont {Anguelov}, \citenamefont
  {Erhan}, \citenamefont {Vanhoucke},\ and\ \citenamefont
  {Rabinovich}}]{szegedy2015going}%
  \BibitemOpen
  \bibfield  {author} {\bibinfo {author} {\bibfnamefont {C.}~\bibnamefont
  {Szegedy}}, \bibinfo {author} {\bibfnamefont {W.}~\bibnamefont {Liu}},
  \bibinfo {author} {\bibfnamefont {Y.}~\bibnamefont {Jia}}, \bibinfo {author}
  {\bibfnamefont {P.}~\bibnamefont {Sermanet}}, \bibinfo {author}
  {\bibfnamefont {S.}~\bibnamefont {Reed}}, \bibinfo {author} {\bibfnamefont
  {D.}~\bibnamefont {Anguelov}}, \bibinfo {author} {\bibfnamefont
  {D.}~\bibnamefont {Erhan}}, \bibinfo {author} {\bibfnamefont
  {V.}~\bibnamefont {Vanhoucke}}, \ and\ \bibinfo {author} {\bibfnamefont
  {A.}~\bibnamefont {Rabinovich}},\ }\bibfield  {title} {\enquote {\bibinfo
  {title} {Going deeper with convolutions},}\ }in\ \href@noop {} {\emph
  {\bibinfo {booktitle} {Proceedings of the IEEE Conference on Computer Vision
  and Pattern Recognition}}}\ (\bibinfo {year} {2015})\ pp.\ \bibinfo {pages}
  {1--9}\BibitemShut {NoStop}%
\bibitem [{\citenamefont {Kamnitsas}\ \emph {et~al.}(2016)\citenamefont
  {Kamnitsas}, \citenamefont {Ledig}, \citenamefont {Newcombe}, \citenamefont
  {Simpson}, \citenamefont {Kane}, \citenamefont {Menon}, \citenamefont
  {Rueckert},\ and\ \citenamefont {Glocker}}]{kamnitsas2016efficient}%
  \BibitemOpen
  \bibfield  {author} {\bibinfo {author} {\bibfnamefont {K.}~\bibnamefont
  {Kamnitsas}}, \bibinfo {author} {\bibfnamefont {C.}~\bibnamefont {Ledig}},
  \bibinfo {author} {\bibfnamefont {V.~F.}\ \bibnamefont {Newcombe}}, \bibinfo
  {author} {\bibfnamefont {J.~P.}\ \bibnamefont {Simpson}}, \bibinfo {author}
  {\bibfnamefont {A.~D.}\ \bibnamefont {Kane}}, \bibinfo {author}
  {\bibfnamefont {D.~K.}\ \bibnamefont {Menon}}, \bibinfo {author}
  {\bibfnamefont {D.}~\bibnamefont {Rueckert}}, \ and\ \bibinfo {author}
  {\bibfnamefont {B.}~\bibnamefont {Glocker}},\ }\bibfield  {title} {\enquote
  {\bibinfo {title} {Efficient multi-scale 3d cnn with fully connected crf for
  accurate brain lesion segmentation},}\ }\href@noop {} {\bibfield  {journal}
  {\bibinfo  {journal} {arXiv preprint arXiv:1603.05959}\ } (\bibinfo {year}
  {2016})}\BibitemShut {NoStop}%
\bibitem [{\citenamefont {Dolz}, \citenamefont {Desrosiers},\ and\
  \citenamefont {Ayed}(2016)}]{dolz20163d}%
  \BibitemOpen
  \bibfield  {author} {\bibinfo {author} {\bibfnamefont {J.}~\bibnamefont
  {Dolz}}, \bibinfo {author} {\bibfnamefont {C.}~\bibnamefont {Desrosiers}}, \
  and\ \bibinfo {author} {\bibfnamefont {I.~B.}\ \bibnamefont {Ayed}},\
  }\bibfield  {title} {\enquote {\bibinfo {title} {3d fully convolutional
  networks for subcortical segmentation in mri: A large-scale study},}\
  }\href@noop {} {\bibfield  {journal} {\bibinfo  {journal} {arXiv preprint
  arXiv:1612.03925}\ } (\bibinfo {year} {2016})}\BibitemShut {NoStop}%
\bibitem [{\citenamefont {Shakeri}\ \emph {et~al.}(2016)\citenamefont
  {Shakeri}, \citenamefont {Tsogkas}, \citenamefont {Ferrante}, \citenamefont
  {Lippe}, \citenamefont {Kadoury}, \citenamefont {Paragios},\ and\
  \citenamefont {Kokkinos}}]{shakeri2016sub}%
  \BibitemOpen
  \bibfield  {author} {\bibinfo {author} {\bibfnamefont {M.}~\bibnamefont
  {Shakeri}}, \bibinfo {author} {\bibfnamefont {S.}~\bibnamefont {Tsogkas}},
  \bibinfo {author} {\bibfnamefont {E.}~\bibnamefont {Ferrante}}, \bibinfo
  {author} {\bibfnamefont {S.}~\bibnamefont {Lippe}}, \bibinfo {author}
  {\bibfnamefont {S.}~\bibnamefont {Kadoury}}, \bibinfo {author} {\bibfnamefont
  {N.}~\bibnamefont {Paragios}}, \ and\ \bibinfo {author} {\bibfnamefont
  {I.}~\bibnamefont {Kokkinos}},\ }\bibfield  {title} {\enquote {\bibinfo
  {title} {Sub-cortical brain structure segmentation using f-cnn's},}\ }in\
  \href@noop {} {\emph {\bibinfo {booktitle} {Biomedical Imaging (ISBI), 2016
  IEEE 13th International Symposium on}}}\ (\bibinfo {organization} {IEEE},\
  \bibinfo {year} {2016})\ pp.\ \bibinfo {pages} {269--272}\BibitemShut
  {NoStop}%
\bibitem [{\citenamefont {Cha}\ \emph {et~al.}(2016)\citenamefont {Cha},
  \citenamefont {Hadjiiski}, \citenamefont {Samala}, \citenamefont {Chan},
  \citenamefont {Caoili},\ and\ \citenamefont {Cohan}}]{cha2016urinary}%
  \BibitemOpen
  \bibfield  {author} {\bibinfo {author} {\bibfnamefont {K.~H.}\ \bibnamefont
  {Cha}}, \bibinfo {author} {\bibfnamefont {L.}~\bibnamefont {Hadjiiski}},
  \bibinfo {author} {\bibfnamefont {R.~K.}\ \bibnamefont {Samala}}, \bibinfo
  {author} {\bibfnamefont {H.-P.}\ \bibnamefont {Chan}}, \bibinfo {author}
  {\bibfnamefont {E.~M.}\ \bibnamefont {Caoili}}, \ and\ \bibinfo {author}
  {\bibfnamefont {R.~H.}\ \bibnamefont {Cohan}},\ }\bibfield  {title} {\enquote
  {\bibinfo {title} {Urinary bladder segmentation in ct urography using
  deep-learning convolutional neural network and level sets},}\ }\href@noop {}
  {\bibfield  {journal} {\bibinfo  {journal} {Medical physics}\ }\textbf
  {\bibinfo {volume} {43}},\ \bibinfo {pages} {1882--1896} (\bibinfo {year}
  {2016})}\BibitemShut {NoStop}%
\bibitem [{\citenamefont {Grady}(2005)}]{gradyML}%
  \BibitemOpen
  \bibfield  {author} {\bibinfo {author} {\bibfnamefont {L.}~\bibnamefont
  {Grady}},\ }\bibfield  {title} {\enquote {\bibinfo {title} {Multilabel random
  walker image segmentation using prior models},}\ }in\ \href@noop {} {\emph
  {\bibinfo {booktitle} {Computer Vision and Pattern Recognition, 2005. CVPR
  2005. IEEE Computer Society Conference on}}},\ Vol.~\bibinfo {volume} {1}\
  (\bibinfo {organization} {IEEE},\ \bibinfo {year} {2005})\ pp.\ \bibinfo
  {pages} {763--770}\BibitemShut {NoStop}%
\bibitem [{\citenamefont {He}\ \emph {et~al.}(2015)\citenamefont {He},
  \citenamefont {Zhang}, \citenamefont {Ren},\ and\ \citenamefont
  {Sun}}]{he2015delving}%
  \BibitemOpen
  \bibfield  {author} {\bibinfo {author} {\bibfnamefont {K.}~\bibnamefont
  {He}}, \bibinfo {author} {\bibfnamefont {X.}~\bibnamefont {Zhang}}, \bibinfo
  {author} {\bibfnamefont {S.}~\bibnamefont {Ren}}, \ and\ \bibinfo {author}
  {\bibfnamefont {J.}~\bibnamefont {Sun}},\ }\bibfield  {title} {\enquote
  {\bibinfo {title} {Delving deep into rectifiers: Surpassing human-level
  performance on imagenet classification},}\ }in\ \href@noop {} {\emph
  {\bibinfo {booktitle} {Proceedings of the IEEE International Conference on
  Computer Vision}}}\ (\bibinfo {year} {2015})\ pp.\ \bibinfo {pages}
  {1026--1034}\BibitemShut {NoStop}%
\bibitem [{\citenamefont {Tieleman}\ and\ \citenamefont
  {Hinton}(2012)}]{tieleman2012lecture}%
  \BibitemOpen
  \bibfield  {author} {\bibinfo {author} {\bibfnamefont {T.}~\bibnamefont
  {Tieleman}}\ and\ \bibinfo {author} {\bibfnamefont {G.}~\bibnamefont
  {Hinton}},\ }\bibfield  {title} {\enquote {\bibinfo {title} {Lecture
  6.5-rmsprop: Divide the gradient by a running average of its recent
  magnitude},}\ }\href@noop {} {\bibfield  {journal} {\bibinfo  {journal}
  {COURSERA: Neural networks for machine learning}\ }\textbf {\bibinfo {volume}
  {4}} (\bibinfo {year} {2012})}\BibitemShut {NoStop}%
\bibitem [{\citenamefont {Bergstra}\ \emph {et~al.}(2010)\citenamefont
  {Bergstra}, \citenamefont {Breuleux}, \citenamefont {Bastien}, \citenamefont
  {Lamblin}, \citenamefont {Pascanu}, \citenamefont {Desjardins}, \citenamefont
  {Turian}, \citenamefont {Warde-Farley},\ and\ \citenamefont
  {Bengio}}]{bergstra2010theano}%
  \BibitemOpen
  \bibfield  {author} {\bibinfo {author} {\bibfnamefont {J.}~\bibnamefont
  {Bergstra}}, \bibinfo {author} {\bibfnamefont {O.}~\bibnamefont {Breuleux}},
  \bibinfo {author} {\bibfnamefont {F.}~\bibnamefont {Bastien}}, \bibinfo
  {author} {\bibfnamefont {P.}~\bibnamefont {Lamblin}}, \bibinfo {author}
  {\bibfnamefont {R.}~\bibnamefont {Pascanu}}, \bibinfo {author} {\bibfnamefont
  {G.}~\bibnamefont {Desjardins}}, \bibinfo {author} {\bibfnamefont
  {J.}~\bibnamefont {Turian}}, \bibinfo {author} {\bibfnamefont
  {D.}~\bibnamefont {Warde-Farley}}, \ and\ \bibinfo {author} {\bibfnamefont
  {Y.}~\bibnamefont {Bengio}},\ }\bibfield  {title} {\enquote {\bibinfo {title}
  {Theano: A cpu and gpu math compiler in python},}\ }in\ \href@noop {} {\emph
  {\bibinfo {booktitle} {Proc. 9th Python in Science Conf}}}\ (\bibinfo {year}
  {2010})\ pp.\ \bibinfo {pages} {1--7}\BibitemShut {NoStop}%
\bibitem [{\citenamefont {Kass}, \citenamefont {Witkin},\ and\ \citenamefont
  {Terzopoulos}(1988)}]{Kass1988}%
  \BibitemOpen
  \bibfield  {author} {\bibinfo {author} {\bibfnamefont {M.}~\bibnamefont
  {Kass}}, \bibinfo {author} {\bibfnamefont {A.}~\bibnamefont {Witkin}}, \ and\
  \bibinfo {author} {\bibfnamefont {D.}~\bibnamefont {Terzopoulos}},\
  }\bibfield  {title} {\enquote {\bibinfo {title} {Snakes: Active contour
  models},}\ }\href {\doibase 10.1007/BF00133570} {\bibfield  {journal}
  {\bibinfo  {journal} {International Journal of Computer Vision}\ }\textbf
  {\bibinfo {volume} {1}},\ \bibinfo {pages} {321--331} (\bibinfo {year}
  {1988})}\BibitemShut {NoStop}%
\bibitem [{\citenamefont {Oezcelik}\ and\ \citenamefont
  {DeMeester}(2011)}]{pmid21477778}%
  \BibitemOpen
  \bibfield  {author} {\bibinfo {author} {\bibfnamefont {A.}~\bibnamefont
  {Oezcelik}}\ and\ \bibinfo {author} {\bibfnamefont {S.~R.}\ \bibnamefont
  {DeMeester}},\ }\bibfield  {title} {\enquote {\bibinfo {title} {{{G}eneral
  anatomy of the esophagus}},}\ }\href@noop {} {\bibfield  {journal} {\bibinfo
  {journal} {Thorac Surg Clin}\ }\textbf {\bibinfo {volume} {21}},\ \bibinfo
  {pages} {289--297} (\bibinfo {year} {2011})}\BibitemShut {NoStop}%
\bibitem [{\citenamefont {Grady}(2006)}]{gradyRW}%
  \BibitemOpen
  \bibfield  {author} {\bibinfo {author} {\bibfnamefont {L.}~\bibnamefont
  {Grady}},\ }\bibfield  {title} {\enquote {\bibinfo {title} {Random walks for
  image segmentation},}\ }\href@noop {} {\bibfield  {journal} {\bibinfo
  {journal} {IEEE transactions on pattern analysis and machine intelligence}\
  }\textbf {\bibinfo {volume} {28}},\ \bibinfo {pages} {1768--1783} (\bibinfo
  {year} {2006})}\BibitemShut {NoStop}%
\bibitem [{\citenamefont {van~der Walt}\ \emph {et~al.}(2014)\citenamefont
  {van~der Walt}, \citenamefont {{S}ch\"onberger}, \citenamefont
  {{Nunez-Iglesias}}, \citenamefont {{B}oulogne}, \citenamefont {{W}arner},
  \citenamefont {{Y}ager}, \citenamefont {{G}ouillart}, \citenamefont {{Y}u},\
  and\ \citenamefont {the scikit-image contributors}}]{scikit-image}%
  \BibitemOpen
  \bibfield  {author} {\bibinfo {author} {\bibfnamefont {S.}~\bibnamefont
  {van~der Walt}}, \bibinfo {author} {\bibfnamefont {J.~L.}\ \bibnamefont
  {{S}ch\"onberger}}, \bibinfo {author} {\bibfnamefont {J.}~\bibnamefont
  {{Nunez-Iglesias}}}, \bibinfo {author} {\bibfnamefont {F.}~\bibnamefont
  {{B}oulogne}}, \bibinfo {author} {\bibfnamefont {J.~D.}\ \bibnamefont
  {{W}arner}}, \bibinfo {author} {\bibfnamefont {N.}~\bibnamefont {{Y}ager}},
  \bibinfo {author} {\bibfnamefont {E.}~\bibnamefont {{G}ouillart}}, \bibinfo
  {author} {\bibfnamefont {T.}~\bibnamefont {{Y}u}}, \ and\ \bibinfo {author}
  {\bibnamefont {the scikit-image contributors}},\ }\bibfield  {title}
  {\enquote {\bibinfo {title} {scikit-image: image processing in {P}ython},}\
  }\href {\doibase 10.7717/peerj.453} {\bibfield  {journal} {\bibinfo
  {journal} {PeerJ}\ }\textbf {\bibinfo {volume} {2}},\ \bibinfo {pages} {e453}
  (\bibinfo {year} {2014})}\BibitemShut {NoStop}%
\bibitem [{\citenamefont {Adebahr}\ \emph {et~al.}(2015)\citenamefont
  {Adebahr}, \citenamefont {Collette}, \citenamefont {Shash}, \citenamefont
  {Lambrecht}, \citenamefont {Le~Pechoux}, \citenamefont {Faivre-Finn},
  \citenamefont {De~Ruysscher}, \citenamefont {Peulen}, \citenamefont
  {Belderbos}, \citenamefont {Dziadziuszko} \emph
  {et~al.}}]{adebahr2015lungtech}%
  \BibitemOpen
  \bibfield  {author} {\bibinfo {author} {\bibfnamefont {S.}~\bibnamefont
  {Adebahr}}, \bibinfo {author} {\bibfnamefont {S.}~\bibnamefont {Collette}},
  \bibinfo {author} {\bibfnamefont {E.}~\bibnamefont {Shash}}, \bibinfo
  {author} {\bibfnamefont {M.}~\bibnamefont {Lambrecht}}, \bibinfo {author}
  {\bibfnamefont {C.}~\bibnamefont {Le~Pechoux}}, \bibinfo {author}
  {\bibfnamefont {C.}~\bibnamefont {Faivre-Finn}}, \bibinfo {author}
  {\bibfnamefont {D.}~\bibnamefont {De~Ruysscher}}, \bibinfo {author}
  {\bibfnamefont {H.}~\bibnamefont {Peulen}}, \bibinfo {author} {\bibfnamefont
  {J.}~\bibnamefont {Belderbos}}, \bibinfo {author} {\bibfnamefont
  {R.}~\bibnamefont {Dziadziuszko}},  \emph {et~al.},\ }\bibfield  {title}
  {\enquote {\bibinfo {title} {Lungtech, an eortc phase ii trial of
  stereotactic body radiotherapy for centrally located lung tumours: a clinical
  perspective},}\ }\href@noop {} {\bibfield  {journal} {\bibinfo  {journal}
  {The British journal of radiology}\ }\textbf {\bibinfo {volume} {88}},\
  \bibinfo {pages} {20150036} (\bibinfo {year} {2015})}\BibitemShut {NoStop}%
\bibitem [{Note1()}]{Note1}%
  \BibitemOpen
  \bibinfo {note}
  {Https://www.synapse.org/$\#$!Synapse:syn3193805/wiki/89480}\BibitemShut
  {NoStop}%
\bibitem [{\citenamefont {S{\o}rensen}(1948)}]{sorensen1948method}%
  \BibitemOpen
  \bibfield  {author} {\bibinfo {author} {\bibfnamefont {T.}~\bibnamefont
  {S{\o}rensen}},\ }\bibfield  {title} {\enquote {\bibinfo {title} {A method of
  establishing groups of equal amplitude in plant sociology based on similarity
  of species and its application to analyses of the vegetation on danish
  commons},}\ }\href@noop {} {\bibfield  {journal} {\bibinfo  {journal} {Biol.
  Skr.}\ }\textbf {\bibinfo {volume} {5}},\ \bibinfo {pages} {1--34} (\bibinfo
  {year} {1948})}\BibitemShut {NoStop}%
\bibitem [{\citenamefont {Feulner}\ \emph {et~al.}(2011)\citenamefont
  {Feulner}, \citenamefont {Zhou}, \citenamefont {Hammon}, \citenamefont
  {Seifert}, \citenamefont {Huber}, \citenamefont {Comaniciu}, \citenamefont
  {Hornegger},\ and\ \citenamefont {Cavallaro}}]{feulner2011probabilistic}%
  \BibitemOpen
  \bibfield  {author} {\bibinfo {author} {\bibfnamefont {J.}~\bibnamefont
  {Feulner}}, \bibinfo {author} {\bibfnamefont {S.~K.}\ \bibnamefont {Zhou}},
  \bibinfo {author} {\bibfnamefont {M.}~\bibnamefont {Hammon}}, \bibinfo
  {author} {\bibfnamefont {S.}~\bibnamefont {Seifert}}, \bibinfo {author}
  {\bibfnamefont {M.}~\bibnamefont {Huber}}, \bibinfo {author} {\bibfnamefont
  {D.}~\bibnamefont {Comaniciu}}, \bibinfo {author} {\bibfnamefont
  {J.}~\bibnamefont {Hornegger}}, \ and\ \bibinfo {author} {\bibfnamefont
  {A.}~\bibnamefont {Cavallaro}},\ }\bibfield  {title} {\enquote {\bibinfo
  {title} {A probabilistic model for automatic segmentation of the esophagus in
  3-d ct scans},}\ }\href@noop {} {\bibfield  {journal} {\bibinfo  {journal}
  {IEEE transactions on medical imaging}\ }\textbf {\bibinfo {volume} {30}},\
  \bibinfo {pages} {1252--1264} (\bibinfo {year} {2011})}\BibitemShut {NoStop}%
\bibitem [{\citenamefont {Cohen}\ \emph {et~al.}(2010)\citenamefont {Cohen},
  \citenamefont {Paskalev}, \citenamefont {Litwin}, \citenamefont {Price},
  \citenamefont {Feigenberg},\ and\ \citenamefont
  {Konski}}]{cohenEsophagusMotion}%
  \BibitemOpen
  \bibfield  {author} {\bibinfo {author} {\bibfnamefont {R.~J.}\ \bibnamefont
  {Cohen}}, \bibinfo {author} {\bibfnamefont {K.}~\bibnamefont {Paskalev}},
  \bibinfo {author} {\bibfnamefont {S.}~\bibnamefont {Litwin}}, \bibinfo
  {author} {\bibfnamefont {R.~A.}\ \bibnamefont {Price}}, \bibinfo {author}
  {\bibfnamefont {S.~J.}\ \bibnamefont {Feigenberg}}, \ and\ \bibinfo {author}
  {\bibfnamefont {A.~A.}\ \bibnamefont {Konski}},\ }\bibfield  {title}
  {\enquote {\bibinfo {title} {{{E}sophageal motion during radiotherapy:
  quantification and margin implications}},}\ }\href@noop {} {\bibfield
  {journal} {\bibinfo  {journal} {Dis. Esophagus}\ }\textbf {\bibinfo {volume}
  {23}},\ \bibinfo {pages} {473--479} (\bibinfo {year} {2010})}\BibitemShut
  {NoStop}%
\bibitem [{\citenamefont {Palmer}\ \emph {et~al.}(2014)\citenamefont {Palmer},
  \citenamefont {Yang}, \citenamefont {Pan},\ and\ \citenamefont
  {Court}}]{palmerEsophagusMotionCardiac}%
  \BibitemOpen
  \bibfield  {author} {\bibinfo {author} {\bibfnamefont {J.}~\bibnamefont
  {Palmer}}, \bibinfo {author} {\bibfnamefont {J.}~\bibnamefont {Yang}},
  \bibinfo {author} {\bibfnamefont {T.}~\bibnamefont {Pan}}, \ and\ \bibinfo
  {author} {\bibfnamefont {L.~E.}\ \bibnamefont {Court}},\ }\bibfield  {title}
  {\enquote {\bibinfo {title} {{{M}otion of the esophagus due to cardiac
  motion}},}\ }\href@noop {} {\bibfield  {journal} {\bibinfo  {journal} {PLoS
  ONE}\ }\textbf {\bibinfo {volume} {9}},\ \bibinfo {pages} {e89126} (\bibinfo
  {year} {2014})}\BibitemShut {NoStop}%
\bibitem [{\citenamefont {Yamashita}\ \emph {et~al.}(2010)\citenamefont
  {Yamashita}, \citenamefont {Haga}, \citenamefont {Hayakawa}, \citenamefont
  {Okuma}, \citenamefont {Yoda}, \citenamefont {Okano}, \citenamefont {Tanaka},
  \citenamefont {Imae}, \citenamefont {Ohtomo},\ and\ \citenamefont
  {Nakagawa}}]{yamashitaEsophagusMotionCBCT}%
  \BibitemOpen
  \bibfield  {author} {\bibinfo {author} {\bibfnamefont {H.}~\bibnamefont
  {Yamashita}}, \bibinfo {author} {\bibfnamefont {A.}~\bibnamefont {Haga}},
  \bibinfo {author} {\bibfnamefont {Y.}~\bibnamefont {Hayakawa}}, \bibinfo
  {author} {\bibfnamefont {K.}~\bibnamefont {Okuma}}, \bibinfo {author}
  {\bibfnamefont {K.}~\bibnamefont {Yoda}}, \bibinfo {author} {\bibfnamefont
  {Y.}~\bibnamefont {Okano}}, \bibinfo {author} {\bibfnamefont
  {K.}~\bibnamefont {Tanaka}}, \bibinfo {author} {\bibfnamefont
  {T.}~\bibnamefont {Imae}}, \bibinfo {author} {\bibfnamefont {K.}~\bibnamefont
  {Ohtomo}}, \ and\ \bibinfo {author} {\bibfnamefont {K.}~\bibnamefont
  {Nakagawa}},\ }\bibfield  {title} {\enquote {\bibinfo {title} {{{P}atient
  setup error and day-to-day esophageal motion error analyzed by cone-beam
  computed tomography in radiation therapy}},}\ }\href@noop {} {\bibfield
  {journal} {\bibinfo  {journal} {Acta Oncol}\ }\textbf {\bibinfo {volume}
  {49}},\ \bibinfo {pages} {485--490} (\bibinfo {year} {2010})}\BibitemShut
  {NoStop}%
\end{thebibliography}

%

\end{document}